\documentclass[10pt,a4paper]{article}
\usepackage[T1]{fontenc}
\usepackage[utf8]{inputenc}
\usepackage{newtxtext,newtxmath}
\usepackage[margin=22mm,headheight=14pt,headsep=7mm,footskip=10mm]{geometry}
\usepackage{microtype}
\usepackage{graphicx,xcolor,booktabs,tabularx,array,makecell}
\usepackage{amsmath,mathtools,bm}
\usepackage{algorithm,algpseudocode,float,flafter,placeins,needspace}
\usepackage{changepage,caption,etoolbox,fancyhdr}
\usepackage[numbers,sort&compress]{natbib}
\usepackage{xurl}
\usepackage[colorlinks=true,linkcolor=black,citecolor=black,urlcolor=blue!50!black]{hyperref}
\hypersetup{pdftitle={Learning Complementary Routes for Scouting Persimmon Orchards during Shoot Growth with CoverShare},pdfauthor={Junhao Cao, Tao Wang, Zipeng Xie, Hongyi Xia, Jianian Wu, Lixia Huang},pdfsubject={Preprint},pdfkeywords={persimmon orchard scouting; reinforcement learning; complementary routes; state coverage; credit assignment}}
\newcommand{\ours}[1]{\textbf{#1}}
\newlength{\extralength}
\newlength{\fulllength}
\newcommand{\tablefooter}[1]{\par\vspace{3pt}{\raggedright\footnotesize #1\par}}
\newcommand{\authorcontributions}[1]{\subsection*{Author contributions}#1\par}
\newcommand{\conflictsofinterest}[1]{\subsection*{Conflicts of interest}#1\par}
\newcommand{\dataavailability}[1]{\subsection*{Data and code availability}#1\par}
\newcommand{\funding}[1]{\subsection*{Funding}#1\par}
\begin{document}
\thispagestyle{plain}
\begin{center}
{\Large\bfseries Learning Complementary Routes for Scouting Persimmon Orchards during Shoot Growth with CoverShare\par}
\vspace{10pt}
{\normalsize Junhao Cao\textsuperscript{1}, Tao Wang\textsuperscript{2}, Zipeng Xie\textsuperscript{3}, Hongyi Xia\textsuperscript{1},\\[3pt]
Jianian Wu\textsuperscript{1}, and Lixia Huang\textsuperscript{1,*}\par}
\vspace{8pt}
{\small
\textsuperscript{1}College of Information Engineering, Hunan Applied Technology University,\\
Changde 415100, Hunan, China\\[3pt]
\textsuperscript{2}College of Computer Science and Technology, National University of Defense Technology,\\
Changsha 410073, Hunan, China\\[3pt]
\textsuperscript{3}School of Computer and Software, Nanjing University of Information Science and Technology,\\
Nanjing 210044, Jiangsu, China\par}
\vspace{6pt}
{\small\textsuperscript{*}Correspondence: \href{mailto:Julia2020@hatu.edu.cn}{Julia2020@hatu.edu.cn}\par}
\vspace{5pt}
{\small November 18, 2026\quad\textbullet\quad Preprint\par}
\end{center}
\begin{abstract}
At the study orchard, scouting during persimmon shoot growth focuses on leaf color change, leaf area, and internode length. Several routes may share an access lane, yet each can offer a view of row segments that the others miss. We introduce CoverShare to learn complementary routes under a fixed action budget. Within Policy Gradient for Parallel State Entropy maximization (PGPSE), CoverShare redistributes auxiliary rewards according to the state coverage lost when each policy is omitted and the number of policies visiting each state. It preserves the pooled entropy objective and total auxiliary reward. We simulated the annotated road network of a 6.7-ha site with 107 observable persimmon scouting units across five zones. At 40 actions per route, six distinct policies covered an expected 17.84 units per inspection, compared with 8.28 for six independent runs of one policy, averaged across eight training seeds. With the same policies trained on 40-action routes, extending each inspection route to 80 and 120 actions raised coverage to 22.52 and 28.38 units. Across seven public discrete-state tasks, CoverShare increased state support by a mean of 16.29\% over the team-entropy baseline. Controlled ablations identified leave-one-policy-out coverage as the main contributor to these gains.
\end{abstract}
\medskip

\section{Introduction}\label{sec:introduction}

During shoot growth at the study orchard, scouts examine leaf color change, leaf area, and internode length in different persimmon rows. Two routes starting at the same entrance may pass the same rows before they diverge. A particular row segment can be viewed only from its local observation positions. Given a fixed number of actions per route, how many distinct segments can six routes bring into view, and which segments would be missed if one route were omitted?

The same question arises in reinforcement learning (RL) when several policies explore in parallel under sparse or absent external rewards. Maximum state-entropy exploration addresses this problem by optimizing the state-visitation distribution to encourage broad, balanced coverage \cite{Hazan2019}. Count-based and pseudo-count methods \cite{Bellemare2016}, together with the intrinsic curiosity module (ICM), random network distillation (RND), and ensemble disagreement \cite{Pathak2017,Burda2019,Pathak2019}, provide complementary exploration signals. Most of these methods, however, target a single policy or mutually independent exploration processes. They do not explicitly address redundancy among policies that optimize a shared coverage objective.

Parallel environments and multi-policy training can broaden experience coverage by collecting trajectories simultaneously. Policy Gradient for Parallel State Entropy maximization (PGPSE) extends maximum state-entropy exploration to independently parameterized policies acting in replicated copies of the same Markov decision process \cite{DePaola2025}. Their trajectories define a pooled empirical state distribution, whose entropy serves as a shared team objective. Although pooled entropy favors diverse aggregate coverage, its shared rollout-level scalar does not isolate each policy's marginal contribution or explicitly downweight a redundant trajectory.

The bottleneck-memory task in Fig.~\ref{fig:2} captures a related feature of access between agricultural plots. Farm-road maps connect road segments to parcel entrances \cite{Shan2026Farm}. Multifield routing models likewise restrict entry and exit to designated points \cite{Utamima2023Multi}. Consider several plots served by a single machinery lane running along an irrigation canal. In the benchmark, four rooms play the role of plots, and narrow corridors play the role of access lanes. The topological correspondence concerns which regions a route can reach through a given passage. Two routes can legitimately share that passage. What distinguishes their contributions is whether they reach different ground beyond it.

Existing exploration methods generally construct stronger novelty signals, whereas multi-agent credit assignment methods address cooperative decisions in a shared environment. COMA uses a counterfactual baseline, while VDN and QMIX decompose or factorize joint value functions \cite{Foerster2018,Sunehag2018,Rashid2018}. Shapley-style methods estimate individual contributions from a cooperative-game perspective \cite{Wang2020}. The present setting differs from these shared-environment formulations in two respects. The policies explore replicated environment copies rather than execute joint actions in one world. Their trajectories jointly define a pooled state-entropy objective. Accordingly, the central question is how to allocate auxiliary intrinsic rewards by marginal coverage contribution, rather than decompose a joint action-value function.

\begin{figure}[!htbp]
\centering
\includegraphics[width=.98\linewidth,height=.36\textheight,keepaspectratio]{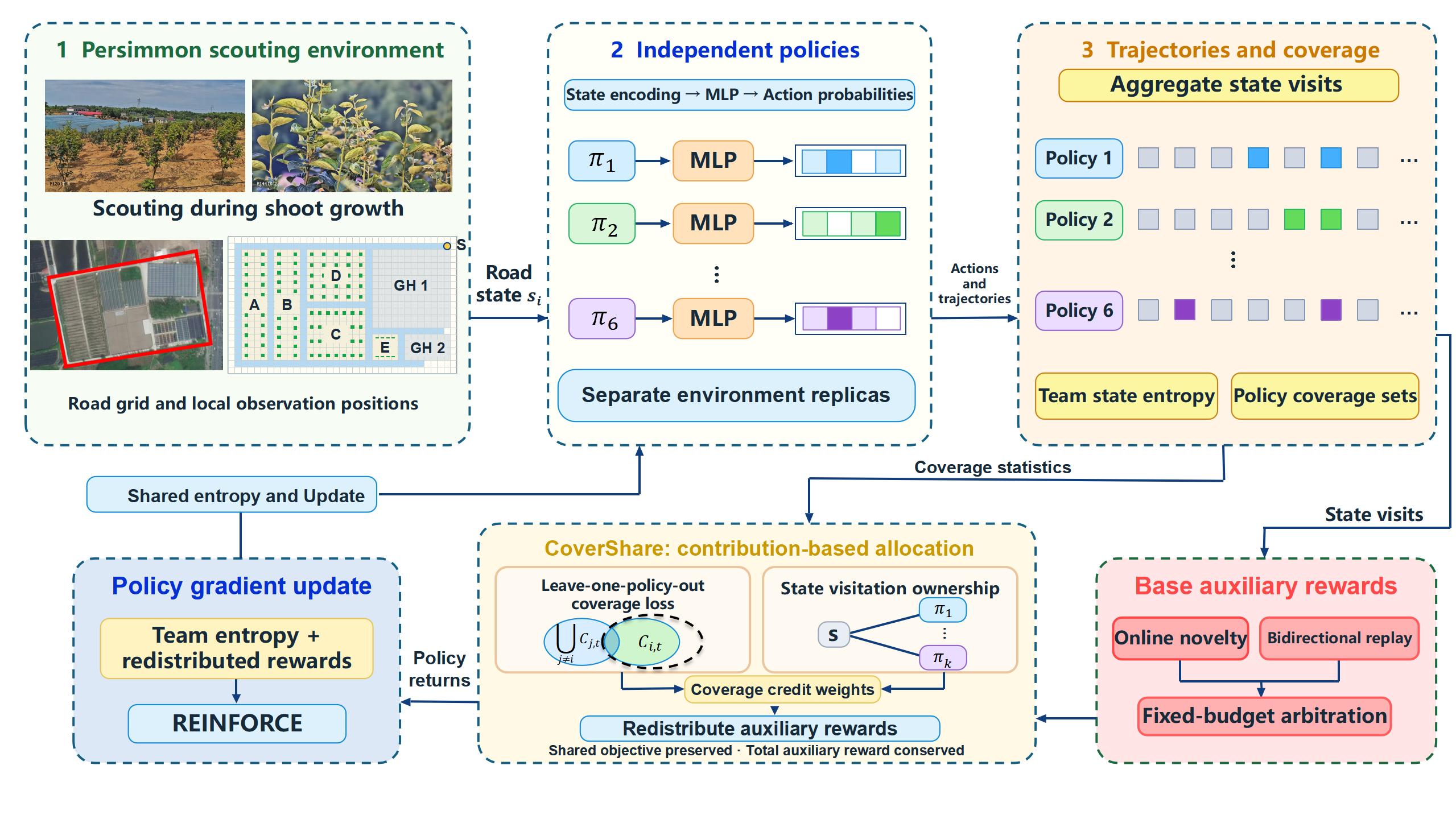}
\caption{Overview of \ours{CoverShare}. At each policy update, the shared team-entropy return is combined with auxiliary rewards redistributed by leave-one-policy-out coverage and state-owner specialization. The redistribution preserves total auxiliary reward mass.}
\label{fig:1}
\end{figure}

We propose CoverShare to resolve this ambiguity. The mechanism combines two complementary sources of policy-specific credit information. First, CoverShare measures the coverage lost when each policy is removed from the pooled trajectory set. This leave-one-policy-out term estimates the policy's contribution to non-redundant team coverage. Second, state-owner specialization augments that trajectory-level contribution to form policy- and transition-specific credit weights. CoverShare neither decomposes nor replaces the original pooled team state-entropy objective. Instead, its weights redistribute non-negative auxiliary intrinsic rewards from the online novelty and replay pathways. Reward-mass renormalization preserves the total auxiliary reward mass. This redistribution favors policies that provide complementary coverage and reduces the signal assigned to redundant trajectories. Fig.~\ref{fig:1} separates this contribution-conditioned pathway from the shared team objective, online novelty, replay, and fixed-budget arbitration.

The main contributions are as follows.

\begin{enumerate}
\item \textbf{Reward allocation by coverage contribution.} CoverShare measures which states would disappear from team coverage if each policy were omitted. It combines this trajectory-level contribution with the number of policies that visited each state to redistribute a fixed auxiliary reward budget while preserving the shared entropy objective.
\item \textbf{Complementary scouting routes.} We evaluate route allocation by the number of distinct persimmon units observable from the simulated routes. At 40 actions per route, six distinct policies covered 9.56 more units on average than six independent runs of one policy.
\item \textbf{Identification of the main source of the gain.} Leave-one-policy-out coverage recovered most of the improvement on both primary tasks. Fixed nonuniform weights and reverse assignment produced results close to the no-credit control, showing that which policy receives the reward matters.
\end{enumerate}

\FloatBarrier
\Needspace{5\baselineskip}
\section{Related methods}\label{sec:related-methods}

\FloatBarrier
\Needspace{3\baselineskip}
\subsection{Exploration and credit assignment}

Exploration methods improve coverage through explicit state-distribution objectives, intrinsic rewards, or behavioral diversity. Maximum state-entropy exploration optimizes the state-visitation distribution, whereas count-based and pseudo-count methods reward rarely visited states. Extensions include hash counts, neural density models, bootstrapped value functions, and parameter-space perturbations \cite{Tang2017,Ostrovski2017,Osband2016,Plappert2018}. ICM uses dynamics-prediction error, RND predicts a fixed random target, and ensemble disagreement estimates model uncertainty. Experience-replay and replay-enhanced methods reuse stored transitions for additional learning \cite{Lin1992,Schaul2016,Mnih2015,Andrychowicz2017,Hessel2018}. Skill-discovery and unsupervised-control methods such as Diversity Is All You Need and non-parametric discriminative rewards encourage diverse or goal-distinguishable behaviors \cite{Eysenbach2019,WardeFarley2019}. These approaches strengthen single-policy exploration or behavioral diversity but do not allocate rewards among policies sharing a coverage objective.

Parallel data collection and multi-policy training mainly improve collection throughput or optimization. Asynchronous reinforcement learning distributes data collection across workers, while population-based training adapts concurrently trained models \cite{Jaderberg2017,Mnih2016}. Policy-gradient and actor-critic algorithms provide optimization machinery across these systems \cite{Sutton2000,Konda1999,Schulman2017,Haarnoja2018,Lillicrap2016,Silver2014}. However, these mechanisms alone do not ensure complementary state coverage across policies. PGPSE, the closest prior method, trains independently parameterized policies in replicated copies of the same Markov decision process. It optimizes the entropy of their pooled empirical state distribution. This design promotes diverse aggregate coverage, but its shared scalar does not identify the marginal contribution of an individual policy.

Cooperative multi-agent reinforcement learning also studies credit assignment under shared objectives. COMA uses a counterfactual baseline, VDN and QMIX decompose team values, and Shapley-style methods estimate marginal contributions. These methods primarily concern agents taking joint actions in a shared environment. By contrast, CoverShare considers policies exploring separate environment copies. It neither learns joint action-values nor decomposes the pooled team state-entropy objective. CoverShare estimates credit from leave-one-policy-out coverage and state-owner specialization, then redistributes non-negative auxiliary intrinsic rewards without changing their total mass.

\FloatBarrier
\Needspace{3\baselineskip}
\subsection{Agricultural coverage path planning}\label{sec:agricultural-coverage-planning}

Mobile crop monitoring can require observations along individual rows, as in greenhouse tomato imaging \cite{Fonteijn2021}. Agricultural coverage planners use field geometry to decide where a vehicle can travel and how it should turn. Pourreau et al. take a field polygon, accessible boundary segments, working width, and minimum turning radius as inputs when generating coverage paths \cite{Pourreau2023Wheeled}. For grain harvesting, N{\o}rremark et al. link crop tracks, headland passes, and feasible turns in a graph that also supports routing unloading vehicles \cite{Norremark2022Harvest}. H{\"o}ffmann et al. optimize the order of field cells and construct smooth paths subject to a curvature bound \cite{Hoffmann2023Curvature}. At the scale of a vineyard, Alshikh Khalil and Palattella connect inspection waypoints across subfields using mapped row alleys, headlands, and crossings between neighboring parcels \cite{Khalil2026Vineyard}.

CoverShare addresses a complementary question during training. Each policy chooses actions from its current discrete state without receiving a global layout or the other policies' routes. The trainer uses common state identifiers across environment copies to compare visitation histories. Discrete transitions omit steering and turning-radius constraints. In a scouting application, marginal credit could help train policies to reach complementary locations, while a vehicle planner would still have to find a drivable path to those locations.

\FloatBarrier
\Needspace{5\baselineskip}
\section{Methods}\label{sec:methods}

\FloatBarrier
\Needspace{3\baselineskip}
\subsection{Problem setting and method overview}\label{sec:problem-setting-and-method-overview}

We consider reward-free exploration in a finite-horizon Markov decision process \(M=(\mathcal{S},\mathcal{A},P,d_0,H)\). Here, \(\mathcal{S}\) and \(\mathcal{A}\) denote the discrete state and action spaces, respectively. The transition function is \(P(s' \mid s,a)\), the initial-state distribution is \(d_0\), and \(H\) is the rollout horizon. The system runs \(N\) independently parameterized policies \(\{\pi_{\theta_i}\}_{i=1}^{N}\) in replicated environments, where \(\theta_i\) parameterizes policy \(i\). These policies neither interact in a shared world nor share parameters. Their trajectories are coupled only through the pooled exploration objective and the coverage-credit calculation.

For rollout group \(b\), the trajectory generated by policy \(i\) is

\begin{equation}
\label{eq:trajectory}
\tau_i^{(b)}
=
\left(
s_{i,0}^{(b)},a_{i,0}^{(b)},s_{i,1}^{(b)},\ldots,
a_{i,H-1}^{(b)},s_{i,H}^{(b)}
\right),
\end{equation}

where \(s_{i,t}^{(b)}\) and \(a_{i,t}^{(b)}\) denote the state and action at time step \(t\), respectively. Pooling the \(N\) trajectories gives

\begin{equation}
\label{eq:pooled-trajectories}
\mathcal{T}^{(b)}=\{\tau_i^{(b)}\}_{i=1}^{N}.
\end{equation}

CoverShare integrates PGPSE's shared team state entropy, each policy's marginal coverage contribution, and reward-conserving redistribution of non-negative auxiliary intrinsic rewards. It retains the shared team-entropy objective and applies coverage credit only to auxiliary rewards from the online novelty and replay pathways. Each discrete state is encoded as a unique one-hot observation. Fig.~\ref{fig:1} shows where the shared objective and policy-specific auxiliary pathway enter the policy update. We omit the rollout-group superscript \(b\) below whenever a single group is understood.

\FloatBarrier
\Needspace{3\baselineskip}
\subsection{Team state-entropy objective}\label{sec:team-state-entropy-objective}

The states visited by all parallel policies define the empirical team state distribution

\begin{equation}
\label{eq:team-state-distribution}
\widehat{d}_{\mathcal{T}^{(b)}}(s)
=
\frac{1}{N(H+1)}
\sum_{i=1}^{N}\sum_{t=0}^{H}
\mathbf{1}\{s_{i,t}^{(b)}=s\},
\end{equation}

where \(\mathbf{1}\{\cdot\}\) is the indicator function. Each trajectory contributes its initial state and \(H\) successor states, giving the denominator \(N(H+1)\). Let \(\mathcal{S}_{\mathrm{valid}}\) denote the environment's valid state set. The team state entropy is

\begin{equation}
\label{eq:team-state-entropy}
\mathcal{H}_{\mathrm{team}}(\mathcal{T}^{(b)})
=
-\sum_{s\in\mathcal{S}_{\mathrm{valid}}}
\widehat{d}_{\mathcal{T}^{(b)}}(s)
\log\widehat{d}_{\mathcal{T}^{(b)}}(s),
\end{equation}

Terms with zero empirical probability are omitted. Team state entropy increases with both the number of visited states and the evenness of visitation. For cross-environment reporting, we normalize this entropy as

\begin{equation}
\label{eq:normalized-coverage-objective}
J_{\mathrm{cov}}(\mathcal{T}^{(b)})
=
\frac{\mathcal{H}_{\mathrm{team}}(\mathcal{T}^{(b)})}
{\log|\mathcal{S}_{\mathrm{valid}}|},
\end{equation}

and the state-support metric is

\begin{equation}
\label{eq:state-support}
\operatorname{Supp}(\mathcal{T}^{(b)})
=
\left|
\left\{
s\in\mathcal{S}_{\mathrm{valid}}
\mid
\widehat{d}_{\mathcal{T}^{(b)}}(s)>0
\right\}
\right|.
\end{equation}

The implementation optimizes the unnormalized \(\mathcal{H}_{\mathrm{team}}\) and uses \(J_{\mathrm{cov}}\) and \(\operatorname{Supp}\) only as reporting metrics. Every policy receives the same team-entropy score, which measures collective exploration but cannot identify the policy responsible for non-redundant coverage.

\FloatBarrier
\Needspace{3\baselineskip}
\subsection{Base auxiliary exploration rewards}\label{sec:base-auxiliary-exploration-rewards}

In the complete method, CoverShare receives non-negative auxiliary rewards from online novelty, bidirectional replay, and fixed-budget arbitration. A discrete state \(s_{i,t}\) is encoded as a \(d\)-dimensional one-hot observation \(x_{i,t}\). For successor state \(s_{i,t+1}\), the within-rollout count novelty is

\begin{equation}
\label{eq:count-novelty}
n_{i,t}
=
\frac{1}
{\sqrt{N_{i,t}^{\mathrm{epi}}(s_{i,t+1})}},
\end{equation}

where \(N_{i,t}^{\mathrm{epi}}(s)\) counts visits by policy \(i\) in the current rollout, including the successor observed at step \(t\). Thus, the first visit gives \(n_{i,t}=1\), and repeated visits progressively reduce the signal.

The online module contains \(K\) forward models. Model \(k\) predicts \(\widehat{x}_{k,i,t+1}\) from \((x_{i,t},a_{i,t})\). Its prediction error, the ensemble-mean error, and ensemble disagreement are

\begin{equation}
\label{eq:online-prediction-errors}
e_{k,i,t}
=
\frac{\left\|
\widehat{x}_{k,i,t+1}-x_{i,t+1}
\right\|_2^2}{d},
\qquad
e_{i,t}
=
\frac{1}{K}\sum_{k=1}^{K}e_{k,i,t},
\end{equation}

\begin{equation}
\label{eq:online-disagreement}
u_{i,t}
=
\frac{1}{d}\sum_{\delta=1}^{d}
\operatorname{Var}_{k}
\left[
\widehat{x}_{k,i,t+1}^{(\delta)}
\right],
\end{equation}

where \(\delta\) indexes observation dimensions and \(\operatorname{Var}_{k}\) denotes population variance across ensemble members. The mean error \(e_{i,t}\) quantifies transition surprise, whereas \(u_{i,t}\) quantifies disagreement among the learned dynamics models. Let \(m_t\) and \(V_t\) denote the running mean and variance of surprise averaged across policies. The adaptive gain is

\begin{equation}
\label{eq:adaptive-gain}
g_{i,t}
=
\sigma\left(
\kappa
\frac{e_{i,t}-m_t}
{\sqrt{V_t}+\epsilon}
\right),
\end{equation}

where \(\sigma(\cdot)\) is the sigmoid function, \(\kappa\) is the gain sensitivity, and \(\epsilon\) is a numerical stabilizer. The online bonus is

\begin{equation}
\label{eq:online-bonus}
b_{i,t}^{\mathrm{on}}
=
\left(
\lambda_n n_{i,t}
+\lambda_u u_{i,t}
\right)
\left(
1+\lambda_g g_{i,t}
\right),
\end{equation}

where \(\lambda_n\), \(\lambda_u\), and \(\lambda_g\) scale count novelty, disagreement, and adaptive gain, respectively. After every transition, the forward ensemble is updated by minimizing mean prediction error across ensemble members and policies.

At rollout termination, the replay module evaluates every stored transition with separate forward and reverse ensembles. The forward models map \((x_{i,t},a_{i,t})\) to \(x_{i,t+1}\), whereas the reverse models map \((x_{i,t+1},a_{i,t})\) to \(x_{i,t}\). The module computes errors \(e_{i,t}^{\mathrm{fwd}}\) and \(e_{i,t}^{\mathrm{rev}}\), together with disagreements \(u_{i,t}^{\mathrm{fwd}}\) and \(u_{i,t}^{\mathrm{rev}}\). The reverse gate is

\begin{equation}
\label{eq:reverse-gate}
\gamma_{i,t}
=
\sigma\left(
\frac{
e_{i,t}^{\mathrm{rev}}
-e_{i,t}^{\mathrm{fwd}}
}{\tau_r}
\right),
\end{equation}

where \(\tau_r\) controls the softness of the forward--reverse mixture. The transition replay value is

\begin{equation}
\label{eq:transition-replay-value}
\begin{aligned}
v_{i,t}^{\mathrm{replay}}
={}&
\gamma_{i,t}
\left(
e_{i,t}^{\mathrm{rev}}+u_{i,t}^{\mathrm{rev}}
\right)\\
&+
\left(1-\gamma_{i,t}\right)
\left(
e_{i,t}^{\mathrm{fwd}}+u_{i,t}^{\mathrm{fwd}}
\right).
\end{aligned}
\end{equation}

For each policy, the replay module selects \(\max\{1,\lfloor Hf_{\mathrm{rep}}\rfloor\}\) transitions with the largest \(v_{i,t}^{\mathrm{replay}}\) to update both model ensembles. The experiments use \(f_{\mathrm{rep}}=0.25\). Selection affects only model training, whereas the replay bonus averages \(v_{i,t}^{\mathrm{replay}}\) over the complete rollout:

\begin{equation}
\label{eq:rollout-replay-bonus}
b_i^{\mathrm{rep}}
=
\frac{\lambda_r}{H}
\sum_{t=0}^{H-1}
v_{i,t}^{\mathrm{replay}},
\end{equation}

where \(\lambda_r\) is the replay-value coefficient.

A meta-controller maintains cross-rollout exponential moving averages of demand and magnitude for the online and replay branches. Online demand combines bonus magnitude with policy-averaged count novelty, disagreement, and adaptive gain. Replay demand combines bonus magnitude with mean forward/reverse model error and replay value. For branch \(m\in\{\mathrm{on},\mathrm{rep}\}\), the budget weight is

\begin{equation}
\label{eq:arbitration-weight}
\omega_i^m
=
\mu_a
+(1-2\mu_a)
\frac{
\exp(\overline{D}_i^m/\tau_a)
}{
\sum_{m'\in\{\mathrm{on},\mathrm{rep}\}}
\exp(\overline{D}_i^{m'}/\tau_a)
},
\end{equation}

Here, \(\overline{D}_i^m\) is the smoothed branch demand, \(\tau_a\) is the arbitration temperature, and \(\mu_a\) is the minimum branch weight. The weights satisfy \(\omega_i^{\mathrm{on}}+\omega_i^{\mathrm{rep}}=1\). The controller divides each branch bonus by its smoothed magnitude before applying the weight. This produces non-negative rewards \(r_{i,t}^{\mathrm{on}}\) and \(r_i^{\mathrm{rep}}\). CoverShare redistributes these rewards but does not create another novelty signal.

\FloatBarrier
\Needspace{3\baselineskip}
\subsection{Marginal coverage credit}\label{sec:marginal-coverage-credit}

The shared team entropy does not reveal which policy supplied non-redundant coverage. CoverShare therefore updates policy-specific coverage sets immediately after each environment transition:

\begin{equation}
\label{eq:coverage-sets}
C_{i,t}
=
\left\{
s_{i,1},s_{i,2},\ldots,s_{i,t+1}
\right\},
\qquad
C_{\mathrm{team},t}
=
\bigcup_{j=1}^{N}C_{j,t}.
\end{equation}

The initial state is included in the team-entropy calculation but not in the CoverShare sets because the implementation begins updating credit after the first transition. The leave-one-policy-out coverage contribution is

\begin{equation}
\label{eq:loo-coverage}
\ell_{i,t}^{\mathrm{loo}}
=
\left|C_{\mathrm{team},t}\right|
-
\left|
\bigcup_{j\neq i}C_{j,t}
\right|.
\end{equation}

If removing policy \(i\) does not change team coverage, then \(\ell_{i,t}^{\mathrm{loo}}=0\): every state found by that policy is replaceable by another policy's coverage. If states disappear from the team set, they are an irreplaceable contribution of policy \(i\). This quantity measures support loss, not the size of \(C_{i,t}\) itself. It is also not an exact marginal difference of the frequency-based entropy objective. The support proxy is deliberate because the targeted failure mode is redundant state discovery; direct marginal entropy is evaluated separately as a mechanism control.

The leave-one-policy-out term summarizes cumulative coverage but does not distinguish transitions within a trajectory. CoverShare therefore counts how many policies own a state:

\begin{equation}
\label{eq:owner-count}
o_t(s)
=
\sum_{j=1}^{N}
\mathbf{1}\{s\in C_{j,t}\},
\end{equation}

and defines the specialization of the current successor state as

\begin{equation}
\label{eq:state-specialization}
q_{i,t}
=
\frac{1}
{o_t(s_{i,t+1})}.
\end{equation}

When only policy \(i\) has visited the current state, \(q_{i,t}=1\); when all \(N\) policies have visited it, \(q_{i,t}=1/N\). Because \(s_{i,t+1}\) has already been inserted into \(C_{i,t}\), its ownership count is at least one, and no stabilizer is required in this denominator.

The raw marginal coverage credit combines the two signals:

\begin{equation}
\label{eq:raw-marginal-credit}
c_{i,t}
=
\alpha_{\mathrm{loo}}\ell_{i,t}^{\mathrm{loo}}
+\alpha_{\mathrm{sp}}q_{i,t},
\end{equation}

where \(\alpha_{\mathrm{loo}}\) and \(\alpha_{\mathrm{sp}}\) control the leave-one-policy-out and specialization terms, respectively. Their default values are 1.0 and 0.5. The first term asks whether the cumulative trajectory is necessary for current team support; the second asks whether the current transition enters a state visited by few policies. Their combination provides both trajectory-level and transition-level coverage credit.

Discrete support changes can make raw credit vary abruptly. The implementation therefore applies a within-rollout exponential moving average:

\begin{equation}
\label{eq:smoothed-credit}
\bar{c}_{i,t}
=
\rho\,\bar{c}_{i,t-1}
+(1-\rho)c_{i,t},
\qquad
\bar{c}_{i,-1}=1,
\end{equation}

where \(\rho\) is the credit-smoothing coefficient and is 0.9 by default. At the beginning of every rollout group, the coverage sets are cleared and the smoothed credits reset to one. The smoothed values are converted to probabilities

\begin{equation}
\label{eq:credit-probability}
p_{i,t}
=
\frac{
\exp(\bar{c}_{i,t}/\tau_c)
}{
\sum_{j=1}^{N}
\exp(\bar{c}_{j,t}/\tau_c)
},
\end{equation}

where the credit temperature \(\tau_c\), 0.5 by default, controls allocation concentration. Smaller values concentrate credit on high-contribution policies, whereas larger values approach a uniform allocation. To retain an auxiliary signal for every policy, CoverShare mixes the softmax probability with a uniform minimum-credit component:

\begin{equation}
\label{eq:covershare-weight}
w_{i,t}^{\mathrm{cov}}
=
N\left[
\frac{\mu}{N}
+(1-\mu)p_{i,t}
\right],
\end{equation}

where \(\mu=0.1\) is the default minimum-credit mixture. This mapping guarantees \(w_{i,t}^{\mathrm{cov}}>0\) and \(\sum_iw_{i,t}^{\mathrm{cov}}=N\).

\FloatBarrier
\Needspace{3\baselineskip}
\subsection{Reward-conserving marginal credit allocation}\label{sec:reward-conserving-marginal-credit-allocation}

Although the CoverShare weights have mean one, multiplying unequal policy rewards by these weights does not by itself preserve their total. To remove reward scale as a competing explanation, define the conservation operator

\begin{equation}
\label{eq:conservation-operator}
\mathcal{C}_i(\mathbf{r},\mathbf{w})
=
w_i r_i
\frac{
\sum_{j=1}^{N}r_j
}{
\sum_{j=1}^{N}w_jr_j
},
\end{equation}

where \(\mathbf{r}=(r_1,\ldots,r_N)\) is a vector of pre-allocation auxiliary rewards and \(\mathbf{w}=(w_1,\ldots,w_N)\) is the corresponding credit-weight vector. Online rewards are redistributed at each transition:

\begin{equation}
\label{eq:allocated-online-reward}
r_{i,t}^{\prime\mathrm{on}}
=
\mathcal{C}_i
\left(
\mathbf{r}_t^{\mathrm{on}},
\mathbf{w}_t^{\mathrm{cov}}
\right).
\end{equation}

The replay module produces one reward per policy at rollout termination, so it uses the final within-rollout weight:

\begin{equation}
\label{eq:allocated-replay-reward}
r_i^{\prime\mathrm{rep}}
=
\mathcal{C}_i
\left(
\mathbf{r}^{\mathrm{rep}},
\mathbf{w}_{H-1}^{\mathrm{cov}}
\right).
\end{equation}

For the non-negative auxiliary rewards evaluated here, whenever the original and weighted totals are positive,

\begin{equation}
\label{eq:reward-conservation}
\sum_{i=1}^{N}r_{i,t}^{\prime\mathrm{on}}
=
\sum_{i=1}^{N}r_{i,t}^{\mathrm{on}},
\qquad
\sum_{i=1}^{N}r_i^{\prime\mathrm{rep}}
=
\sum_{i=1}^{N}r_i^{\mathrm{rep}}.
\end{equation}

Thus, CoverShare changes each policy's share of a fixed auxiliary budget rather than increasing the amount of auxiliary reward. If either total has absolute value below the implementation threshold \(10^{-8}\), or if the totals have opposite signs, the implementation returns the unweighted reward. This fallback is not activated by the evaluated non-negative bonuses except when their total is numerically zero.

For rollout group \(b\), the episodic score for policy \(i\) is

\begin{equation}
\label{eq:episodic-score}
R_i^{(b)}
=
\mathcal{H}_{\mathrm{team}}(\mathcal{T}^{(b)})
+\eta
\left(
\sum_{t=0}^{H-1}
r_{i,t}^{\prime\mathrm{on}(b)}
+r_i^{\prime\mathrm{rep}(b)}
\right),
\end{equation}

where \(\eta\) is the auxiliary intrinsic-reward coefficient. The first term is identical for every policy, whereas the second is contribution-conditioned.

Each parameter update uses \(M_B\) independently sampled rollout groups. Let \(\Theta=\{\theta_i\}_{i=1}^{N}\) denote all policy parameters. The entropy-regularized episodic REINFORCE loss is

\begin{equation}
\label{eq:reinforce-loss}
\begin{aligned}
\mathcal{L}(\Theta)
={}&
-\frac{1}{M_B}
\sum_{b=1}^{M_B}\sum_{i=1}^{N}
R_i^{(b)}
\sum_{t=0}^{H-1}
\log\pi_{\theta_i}
\left(
a_{i,t}^{(b)}
\mid
s_{i,t}^{(b)}
\right)\\
&-
\frac{\beta}{M_BNH}
\sum_{b=1}^{M_B}\sum_{i=1}^{N}\sum_{t=0}^{H-1}
\mathcal{H}
\left(
\pi_{\theta_i}(\cdot\mid s_{i,t}^{(b)})
\right),
\end{aligned}
\end{equation}

where \(\beta\) is the action-entropy coefficient. The entropy regularizer discourages premature policy determinism. The implementation uses no external task reward, discount factor, or learned value baseline; the update follows episodic REINFORCE \cite{SuttonBarto2018,Williams1992}. In the primary method, CoverShare never reweights \(\mathcal{H}_{\mathrm{team}}\). Direct allocation of the team-entropy score is evaluated only as a mechanism-location control.

\FloatBarrier
\Needspace{3\baselineskip}
\subsection{CoverShare algorithm}\label{sec:overall-algorithm}

Algorithm~\ref{alg:covershare-pgpse} summarizes the complete training procedure. Each rollout group resets the environments, episodic counts, replay transitions, coverage sets, and CoverShare smoothing state. The learned dynamics models, running surprise statistics, and meta-controller statistics persist across rollout groups. Online credit is computed after each successor state, whereas team entropy and replay rewards are computed after collecting the complete trajectories.

\begin{algorithm}[H]
\caption{CoverShare training procedure.}
\label{alg:covershare-pgpse}
\fontsize{8}{9.5}\selectfont
\begin{algorithmic}[1]
\Require \(N\) independent policies, \(N\) replicated environments, rollout-group batch size \(M_B\), horizon \(H\), valid state set \(\mathcal{S}_{\mathrm{valid}}\), and online, replay, arbitration, CoverShare, and optimization hyperparameters
\Ensure Updated policy parameters \(\Theta=\{\theta_i\}_{i=1}^{N}\)
\State Initialize policy networks, online forward models, bidirectional replay models, and meta-controller statistics
\For{each training update}
  \State Initialize the batch policy objective and action-entropy statistic
  \For{\(b=1,\ldots,M_B\)}
    \State Reset environments, episodic counts, replay transitions, coverage sets, and CoverShare smoothing state
    \State Observe \(\{s_{i,0}\}_{i=1}^{N}\); initialize trajectory log probabilities and auxiliary returns
    \For{\(t=0,\ldots,H-1\)}
      \State Sample \(a_{i,t}\sim\pi_{\theta_i}(\cdot\mid s_{i,t})\) for every policy and step the environments
      \State Compute count novelty, ensemble error, disagreement, adaptive gain, and \(b_{i,t}^{\mathrm{on}}\)
      \State Update the online forward models and apply the meta-controller to obtain \(r_{i,t}^{\mathrm{on}}\)
      \State Insert \(s_{i,t+1}\) into \(C_{i,t}\) and construct \(C_{\mathrm{team},t}\)
      \State Compute \(\ell_{i,t}^{\mathrm{loo}}\), \(o_t(s_{i,t+1})\), \(q_{i,t}\), and \(c_{i,t}\)
      \State Update \(\bar c_{i,t}\), \(p_{i,t}\), and \(w_{i,t}^{\mathrm{cov}}\)
      \State Redistribute \(r_{i,t}^{\mathrm{on}}\) with \(\mathcal{C}_i\) and accumulate the policy-specific online return
      \State Store successor states, log probabilities, and action entropies
    \EndFor
    \State Compute \(\widehat d_{\mathcal{T}^{(b)}}\), \(\mathcal H_{\mathrm{team}}(\mathcal{T}^{(b)})\), \(J_{\mathrm{cov}}\), and \(\operatorname{Supp}\)
    \State Compute forward/reverse replay values, select the highest-valued transitions for each policy, and update replay models
    \State Arbitrate replay bonuses to obtain \(r_i^{\mathrm{rep}}\)
    \State Redistribute \(r_i^{\mathrm{rep}}\) with \(\mathcal{C}_i\) using \(\mathbf{w}_{H-1}^{\mathrm{cov}}\)
    \State Form \(R_i^{(b)}=\mathcal H_{\mathrm{team}}(\mathcal T^{(b)})+\eta(\text{online return}+r_i^{\prime\mathrm{rep}})\)
    \State Accumulate the policy objective and action entropy for rollout group \(b\)
  \EndFor
  \State Compute \(\mathcal L(\Theta)\), backpropagate, clip policy gradients, and update all policies
\EndFor
\end{algorithmic}
\end{algorithm}

\FloatBarrier
\Needspace{5\baselineskip}
\FloatBarrier
\section{Experiments}\label{sec:experiments-and-results}

We first evaluate how many persimmon scouting units six routes can reach and whether policies trained on short routes can use a longer inspection budget. Controlled tasks isolate the effect of marginal coverage credit. Public benchmarks and original PGPSE protocols assess state coverage across environments, followed by sensitivity and cost analyses.

\FloatBarrier
\Needspace{3\baselineskip}
\subsection{Experimental environments}\label{sec:experimental-environments}

\begin{figure}[!htbp]
\centering\includegraphics[width=.96\linewidth,height=.34\textheight,keepaspectratio]{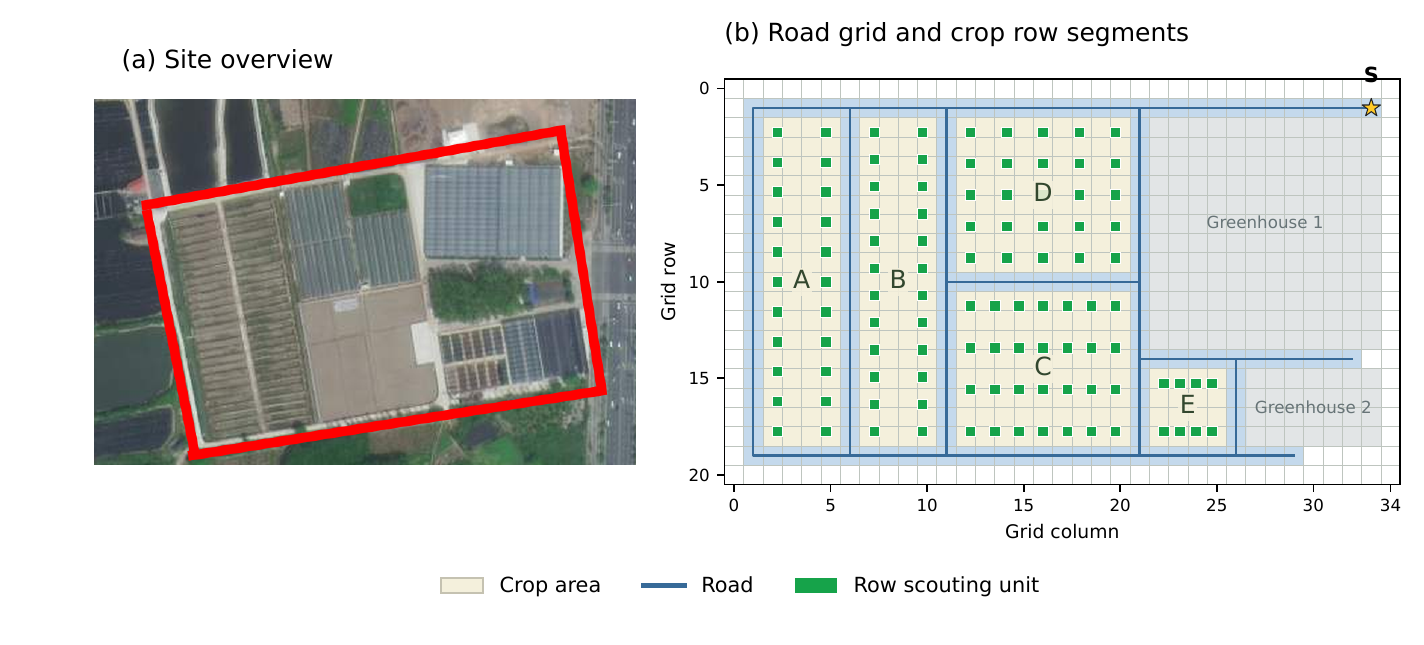}
\caption{Site overview and scouting grid. (a) The red outline marks the study-site boundary. (b) Blue marks roads, cream marks crop zones A--E, gray marks greenhouse areas, and S marks the common entrance. Roads use the simulation grid. Green squares represent 107 scouting units arranged schematically within their zones.}\label{fig:site-network}
\end{figure}

\begin{figure}[!htbp]
\centering\includegraphics[width=.96\linewidth,height=.34\textheight,keepaspectratio]{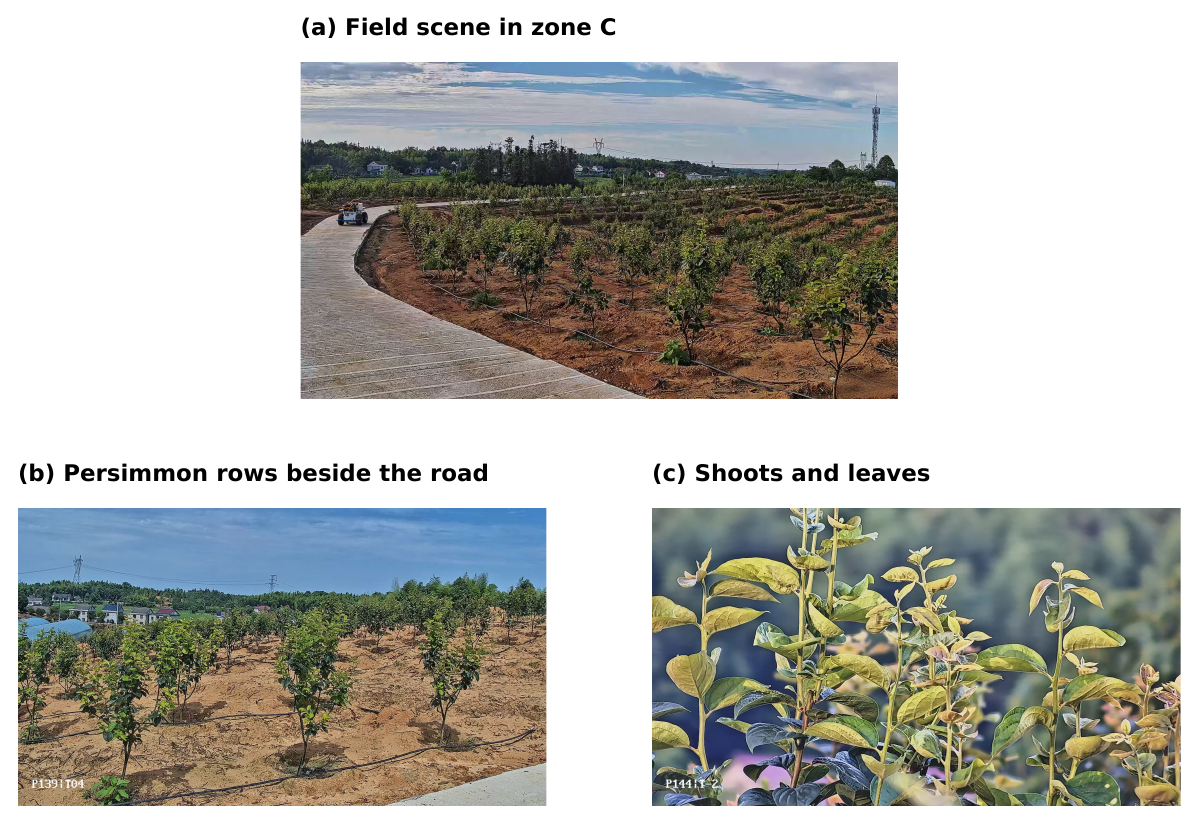}
\caption{Field context for persimmon scouting during shoot growth. (a) Zone C with the inspection platform, road, and planting. (b) Persimmon rows beside a road. (c) Shoots and leaves. Scouting at the site focuses on leaf color change, leaf area, and internode length.}\label{fig:site-context}
\end{figure}

\begin{figure}[!htbp]
\centering\includegraphics[width=.96\linewidth,height=.34\textheight,keepaspectratio]{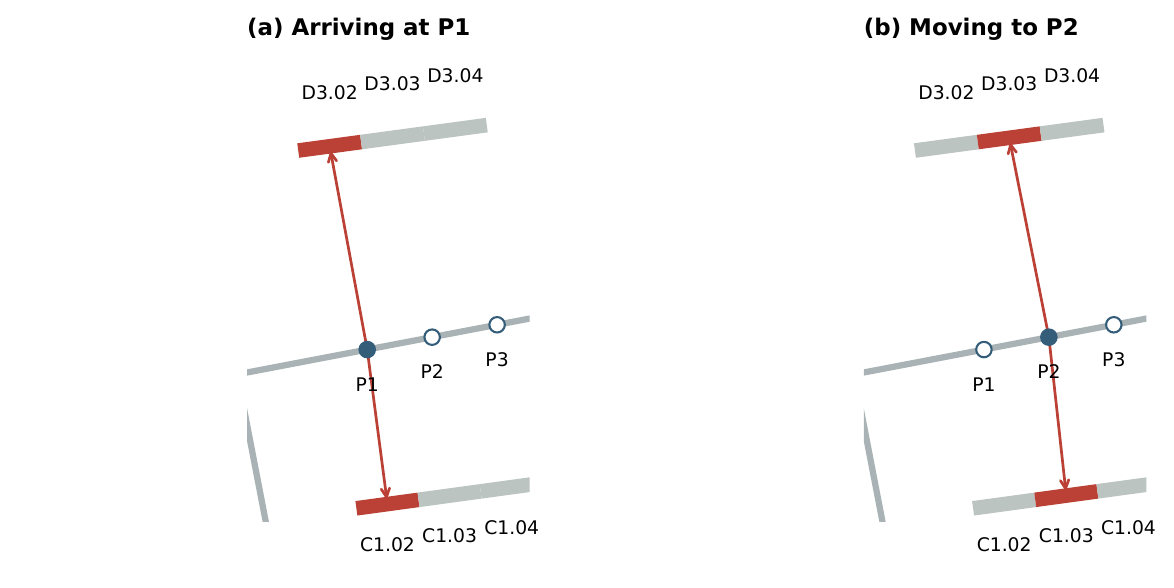}
\caption{Local observation rule. P1--P3 are adjacent road positions between zones C and D. Arrows link each position to row units according to the visibility annotation. Moving from P1 to P2 changes the observable units. Each unit contributes once to team coverage across routes and repeated visits. This local view follows the annotated row geometry.}\label{fig:local-observation}
\end{figure}

Persimmon orchard. We represented the annotated roads of an approximately 6.7-ha site as a four-neighbor grid with 154 reachable positions. All 15 marked target lines in zones A--E correspond to persimmon row segments. We subdivided the lines into locally observable units and merged duplicate units where lines intersected, yielding 107 distinct scouting units. Six routes started at the same entrance. A unit counted once per inspection when any route reached a road position from which that unit was marked as visible. Training credit used road-state visits. Crop visibility annotations were used for evaluation. Figure~\ref{fig:site-network} relates the site overview to the simulation grid. Figure~\ref{fig:site-context} shows the field roads, persimmon rows, and shoots. Figure~\ref{fig:local-observation} illustrates the association between local observation positions and row segments.

\begin{figure}[!htbp]
\centering
\includegraphics[width=.96\linewidth,height=.34\textheight,keepaspectratio]{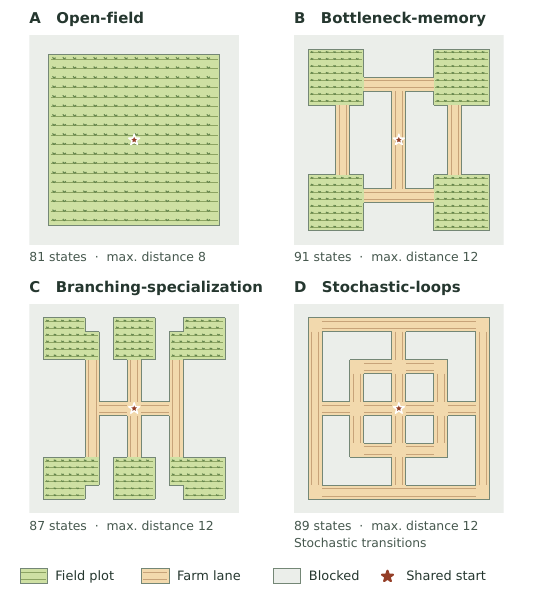}
\caption{Controlled environments: open-field, bottleneck-memory, branching-specialization, and stochastic-loops.}
\label{fig:2}
\end{figure}

The controlled suite comprised four discrete grid-map families (Fig.~\ref{fig:2}). Open-field assessed open coverage, bottleneck-memory assessed constrained passageways, branching-specialization assessed branch specialization, and stochastic-loops assessed stochastic transitions. These structures allowed redundant exploration to be examined under different conditions. In all environments, policies collected trajectories in replicated map copies rather than acting jointly in a shared world.

\textbf{1. Open-field.} Policies start at the center of a deterministic map with an unobstructed interior. This layout provides a simple setting for checking state encoding, rollout collection, and the stability of the auxiliary modules during training.

\textbf{2. Bottleneck-memory.} Bottleneck-memory contains multiple rooms connected by narrow passageways. Repeated traversal of already visited routes consumes opportunities to reach remote rooms, making redundant exploration particularly costly. The environment therefore tests whether parallel policies can reduce repeated bottleneck crossings and direct exploration toward insufficiently covered regions. It is one of the two primary controlled tasks for evaluating marginal coverage credit.

\textbf{3. Branching-specialization.} Branching-specialization connects six outward branches to a central hub, with an explorable region at the end of each branch. Because the branches have similar local structure, independently trained policies can collapse onto the same route. This environment tests whether CoverShare can identify each policy's non-redundant contribution and promote complementary branch coverage. Together with bottleneck-memory, it forms the primary controlled evaluation.

\textbf{4. Stochastic-loops.} The map contains nested loops, intersecting corridors, and alternative routes. Stochastic transitions allow the same action sequence to reach different states, providing a test of coverage under transition noise.

We also reproduced the deterministic Room and stochastic Maze settings from the original PGPSE study. Both maps have 43 reachable states and differ in topology, transition stochasticity, and rollout horizon.

We further assessed cross-environment validity on FrozenLake-v1, Taxi-v3, CliffWalking-v0, and four MiniGrid tasks. The MiniGrid tasks were Empty-8x8-v0, DoorKey-8x8-v0, FourRooms-v0, and LavaGapS7-v0. Gymnasium tasks used native discrete state indices, whereas MiniGrid used an abstraction based on agent position and orientation. All public tasks followed the reward-free state-coverage protocol used for the controlled environments. Accordingly, the outcomes concern normalized team state entropy and state support, not extrinsic return or task success.

\begin{table}[!htbp]
\caption{Evaluation roles of the experimental environments.}\label{tab:1}
\begin{adjustwidth}{-\extralength}{0cm}
\setlength{\linewidth}{\fulllength}

\centering\fontsize{8}{10}\selectfont\setlength{\tabcolsep}{3pt}\renewcommand{\arraystretch}{1.12}
\begin{tabularx}{\fulllength}{@{}p{31mm}p{23mm}X p{35mm}@{}}
\toprule
Environment & Type & Evaluation focus & Role \\
\midrule
Open-field & Controlled & Open-region coverage and implementation stability & Training pipeline check \\
Bottleneck-memory & Controlled & Remote coverage through constrained passageways & Primary task \\
Branching-specialization & Controlled & Complementary exploration across alternative branches & Primary task \\
Stochastic-loops & Controlled & Robustness under stochastic transition loops & Robustness \\
PGPSE Room/Maze & Original protocol & Deterministic Room and stochastic Maze with 43 reachable states each & Full-method replication \\
FrozenLake-v1 & Public benchmark & Discrete exploration under slippery transitions & Cross-environment validation \\
Taxi-v3 & Public benchmark & Discrete pickup-and-delivery state coverage & Cross-environment validation \\
CliffWalking-v0 & Public benchmark & Grid exploration near terminal cliff states & Cross-environment validation \\
MiniGrid tasks & Public benchmark & Empty-8x8-v0, DoorKey-8x8-v0, FourRooms-v0, and LavaGapS7-v0 & Cross-environment validation \\
\bottomrule
\end{tabularx}
\end{adjustwidth}
\end{table}

\FloatBarrier
\Needspace{3\baselineskip}
\subsection{Experimental setup}\label{sec:experimental-setup}

For the orchard simulation, we trained Entropy, Count, Triad, and CoverShare with eight paired seeds and 40 actions per training route. We froze the final policies and evaluated their action distributions at 40, 60, 80, and 120 actions per route. A blocked move consumed one action in place. For each unit, finite-horizon recursion on the grid gave the probability that each policy reached an observing position. Independent route sampling then gave the probability that at least one route reached that unit. Summing these probabilities over the 107 units gave the expected coverage per inspection, which we averaged across training seeds. An internal control ran one CoverShare policy independently six times and averaged uniformly over the six possible choices of policy.

The primary controlled comparison included Entropy, Count, ICM, RND, Online, Triad, and CoverShare. Entropy used only the shared team state-entropy objective. Count, ICM, and RND served as intrinsic-exploration baselines, while Online retained only online novelty. Triad retained online novelty, bidirectional replay, and fixed-budget arbitration but disabled marginal coverage credit. CoverShare used the complete method. All methods trained six independently parameterized policies with matched architectures, budgets, and seeds. Section~\ref{sec:ablation-and-mechanism-analysis} defines the mechanism ablations and credit controls.

The primary controlled comparisons and public benchmarks used eight matched seeds and 300 updates. Each update contained eight rollout groups with a horizon of 20. Policies used two-layer MLPs with 128 hidden units per layer and Adam optimization. Sensitivity and computational-cost analyses instead used five matched seeds. The original PGPSE replication retained linear softmax policies, SGD with exponential decay, five published seeds, and 10,000 updates. Room used 40 rollout groups with a horizon of 8, whereas Maze used 40 groups with a horizon of 10.

The comprehensive comparison added TEE and ETD on the two controlled tasks and MiniGrid-DoorKey-8x8. Trajectory Entropy Exploration (TEE) rewards separation between agents in a learned trajectory representation space \cite{LiZhu2025TEE}. Episodic Novelty through Temporal Distance (ETD) rewards states that are distant from previously visited states within the same episode \cite{Jiang2025ETD}. We integrated each reward into the PGPSE policy update. TEE used the three nearest neighbors among the other five policies at each step. ETD maintained a separate episodic memory for each policy. All three tasks used the same discrete state encoding for every method within a task.

Each adaptation used six independent policies and 288,000 interactions per seed (300 updates, eight rollout groups, and 20 steps per policy). The auxiliary reward coefficient was fixed at 0.1. We used seeds 0--7 and averaged Objective and Support over the final 60 updates for each seed. Within each task, methods matched interaction budgets, state encodings, and seed identifiers.

DoorKey appears in both the comprehensive comparison and the seven-task public suite. Entropy, Triad, and CoverShare use the same experimental records in both presentations; TEE, ETD, and RND provide additional comparators.

For the reinforcement learning benchmarks, the primary metrics were normalized team state entropy (Objective) and state support (Support), each averaged over the final 20\% of training. For 300-update runs, this window contained 60 updates. Learning curves used a centered nine-update moving average for visualization only; final-window statistics used unsmoothed data. Group means are reported with bootstrap 95\% confidence intervals. Matched-seed comparisons used exact two-sided sign-flip tests. Holm correction was applied separately within each environment, metric, and prespecified hypothesis family. Public taskwise comparisons were corrected across seven tasks within each metric.

All runs used Python 3.12.13. The controlled, public, and robustness runs used PyTorch 2.12.0+cu130 with CUDA 13.0. Ablation, protocol, sensitivity, and cost runs used PyTorch 2.5.1 with CUDA 12.4. The computing platform included an Intel Core i9-14900K CPU and an NVIDIA GeForce RTX 3060. TEE, ETD, and the additional DoorKey RND runs used CPU sampling with PyTorch 2.5.1.

\FloatBarrier
\Needspace{3\baselineskip}
\subsection{Experimental results and analysis}\label{sec:experimental-results-and-analysis}

\FloatBarrier
\Needspace{3\baselineskip}
\subsubsection{Persimmon row coverage during shoot-growth scouting}\label{sec:persimmon-results}

\begin{figure}[!htbp]
\centering\includegraphics[width=.96\linewidth,height=.34\textheight,keepaspectratio]{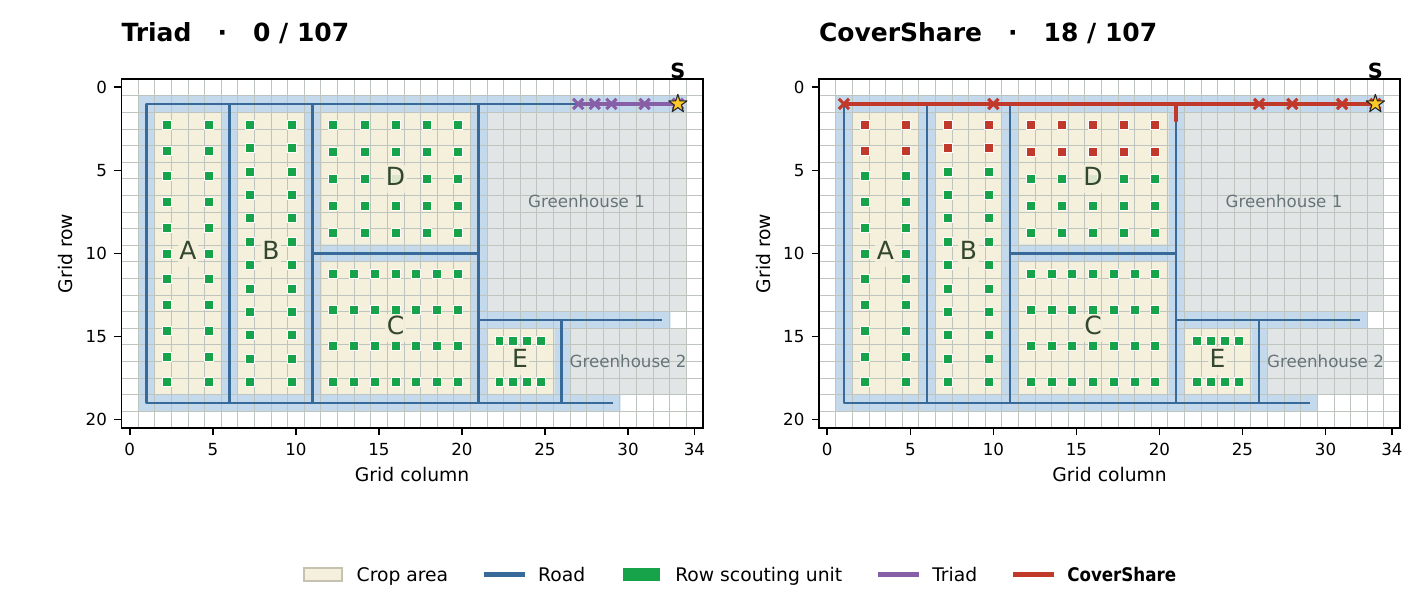}
\caption{One inspection under matched budgets. Both methods use the first rollout with training and evaluation seed 0, six routes, and 40 actions per route. Roads are blue. Purple and red show the respective trajectories and covered units. Crosses mark endpoints. Squares are arranged schematically by zone to display unit membership and coverage. Titles give single-inspection counts; eight-seed means appear in Table~\ref{tab:field-scouting}.}\label{fig:field-scouting}
\end{figure}

At 40 actions per route, six CoverShare policies covered an expected 17.84 persimmon units, compared with 8.28 for six independent runs of one policy (Table~\ref{tab:field-scouting}). The ensemble covered more units in all eight seeds, with a mean gain of 9.56. Zones D, B, and A accounted for 10.25, 4.18, and 3.41 units, respectively. Figure~\ref{fig:field-scouting} compares a single inspection by Triad and CoverShare, which reached 0 and 18 units, respectively.

\begin{table}[!htbp]
\caption{Expected persimmon scouting coverage at different inspection budgets.}\label{tab:field-scouting}
\begin{adjustwidth}{-\extralength}{0cm}
\setlength{\linewidth}{\fulllength}
\centering
\fontsize{8}{10}\selectfont\setlength{\tabcolsep}{4pt}\begin{tabular}{lrrrr}\toprule
Method & 40 & 60 & 80 & 120 \\\midrule
Entropy & 0.22 & 1.29 & 3.18 & 8.32 \\
Count & 0.30 & 1.67 & 3.97 & 9.96 \\
Triad & 0.43 & 2.21 & 4.99 & 11.87 \\
\ours{CoverShare} & \ours{17.84} & \ours{20.12} & \ours{22.52} & \ours{28.38} \\
\midrule
One policy, six runs & 8.28 & 12.78 & 17.12 & 24.96 \\
\bottomrule\end{tabular}\end{adjustwidth}
\tablefooter{Values are expected scouting units covered out of 107, averaged across eight training seeds. Column headings give actions per route. The final row is an internal control.}
\end{table}

\begin{figure}[!htbp]
\centering\includegraphics[width=.96\linewidth,height=.27\textheight,keepaspectratio]{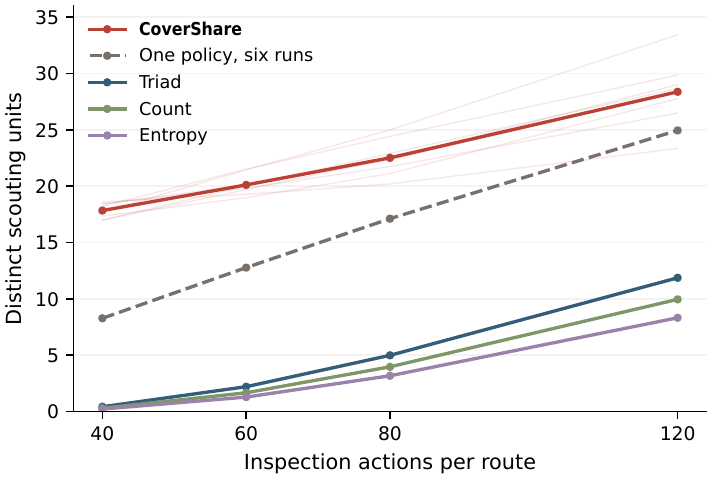}
\caption{Longer inspections with frozen policies trained on 40-action routes. Points show expected coverage averaged across eight training seeds. Faint red lines show individual \ours{CoverShare} seeds. The one-policy control averages six independent runs uniformly over the six choices of policy. The horizontal axis gives evaluation actions per route.}\label{fig:inspection-budget}
\end{figure}

Extending the inspection to twice and three times the training horizon increased coverage by the frozen CoverShare policies to 22.52 and 28.38 units, respectively. Coverage increased from 40 to 80 and again from 80 to 120 actions in all eight seeds. The other methods also covered more units with longer inspections. At 120 actions, six independent runs of one policy covered 24.96 units, narrowing the ensemble advantage from 9.56 to 3.42 units. Figure~\ref{fig:inspection-budget} shows mean coverage at the four budgets and the individual CoverShare seeds.

\FloatBarrier
\Needspace{3\baselineskip}
\subsubsection{Comprehensive comparison of exploration methods}\label{sec:controlled-comparison}\label{sec:trajectory-comparison}\label{sec:cross-environment-comparison}

\begin{figure}[!htbp]
\centering
\includegraphics[width=.96\linewidth,height=.34\textheight,keepaspectratio]{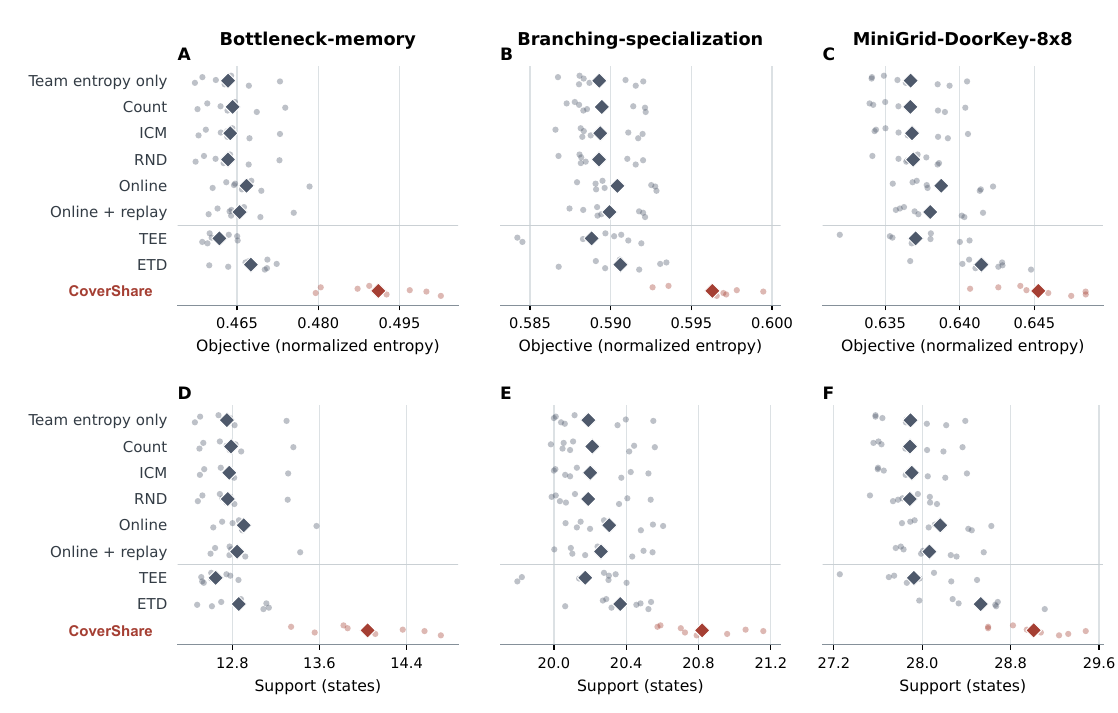}
\caption{Final-window coverage across nine exploration methods. Columns show bottleneck-memory, branching-specialization, and MiniGrid-DoorKey-8x8. Panels A--C show normalized team state entropy (Objective), and D--F show the number of distinct states visited per rollout group (Support). Each dot is one seed's mean over the final 60 updates. Diamonds mark eight-seed means. Team entropy only and Online + replay correspond to Entropy and Triad in Tables~\ref{tab:3} and~\ref{tab:7}.}
\label{fig:trajectory-comparison}
\end{figure}

Figure~\ref{fig:trajectory-comparison} compares final-window performance across nine methods in bottleneck-memory, branching-specialization, and DoorKey. The controlled tasks test repeated coverage along shared corridors and complementary choices at branches, respectively. DoorKey adds discrete interactions with a key and door. We compare final performance by task, then examine training dynamics in Fig.~\ref{fig:controlled-learning}.

\noindent\textbf{Controlled tasks.}\label{sec:performance-in-bottleneck-memory}\label{sec:performance-in-branching-specialization}
In bottleneck-memory, CoverShare covered 14.04 states per rollout group, a 10.13\% increase over Entropy's 12.75 states. Objective increased from 0.4634 to 0.4911, a gain of 5.98\%. Coverage also exceeded ETD's 12.86 states and TEE's 12.65 states. Differences were smaller in branching-specialization, where gains over Entropy were 1.19\% for Objective and 3.11\% for Support. CoverShare covered 0.45 more states than ETD in that task. Both metrics favored CoverShare over ETD in all eight paired seeds on each controlled task.

Among the conventional baselines, Online was closest to CoverShare in final-window means, while Count, ICM, and RND performed similarly to Entropy. On both controlled tasks, Holm-adjusted \(p\) values were 0.0234 against Entropy, Count, and ICM, 0.0078 against Triad, and 0.0156 against RND. Table~\ref{tab:3} provides complete controlled-task means and confidence intervals; Fig.~\ref{fig:trajectory-comparison} shows seed-level results across all nine methods.

\begin{table}[!htbp]
\caption{Final-window performance in controlled environments.}\label{tab:3}
\begin{adjustwidth}{-\extralength}{0cm}
\setlength{\linewidth}{\fulllength}

\centering\fontsize{8}{10}\selectfont\setlength{\tabcolsep}{3pt}\renewcommand{\arraystretch}{1.12}
\begin{tabular*}{\fulllength}{@{\extracolsep{\fill}}llrrrr@{}}
\toprule
Env. & Variant & Objective & Objective 95\% CI & Support & Support 95\% CI \\
\midrule
BM & Entropy & 0.4634 & {[}0.4604, 0.4669{]} & 12.75 & {[}12.60, 12.94{]} \\
BM & Count & 0.4642 & {[}0.4612, 0.4678{]} & 12.79 & {[}12.63, 12.98{]} \\
BM & ICM & 0.4638 & {[}0.4610, 0.4671{]} & 12.77 & {[}12.63, 12.95{]} \\
BM & RND & 0.4633 & {[}0.4605, 0.4668{]} & 12.76 & {[}12.61, 12.94{]} \\
BM & Online & 0.4668 & {[}0.4638, 0.4707{]} & 12.90 & {[}12.76, 13.11{]} \\
BM & Triad & 0.4655 & {[}0.4627, 0.4691{]} & 12.84 & {[}12.71, 13.03{]} \\
BM & \ours{CoverShare} & \ours{0.4911} & \ours{{[}0.4853, 0.4962{]}} & \ours{14.04} & \ours{{[}13.71, 14.33{]}} \\
BS & Entropy & 0.5893 & {[}0.5881, 0.5905{]} & 20.19 & {[}20.06, 20.33{]} \\
BS & Count & 0.5894 & {[}0.5882, 0.5908{]} & 20.21 & {[}20.08, 20.36{]} \\
BS & ICM & 0.5894 & {[}0.5881, 0.5906{]} & 20.20 & {[}20.07, 20.33{]} \\
BS & RND & 0.5893 & {[}0.5880, 0.5905{]} & 20.19 & {[}20.06, 20.33{]} \\
BS & Online & 0.5904 & {[}0.5892, 0.5917{]} & 20.31 & {[}20.18, 20.44{]} \\
BS & Triad & 0.5899 & {[}0.5888, 0.5910{]} & 20.26 & {[}20.13, 20.39{]} \\
BS & \ours{CoverShare} & \ours{0.5963} & \ours{{[}0.5948, 0.5977{]}} & \ours{20.82} & \ours{{[}20.68, 20.96{]}} \\
\bottomrule
\end{tabular*}
\end{adjustwidth}
\tablefooter{Results are averaged across eight seeds. Brackets show bootstrap 95\% confidence intervals. BM: bottleneck-memory; BS: branching-specialization.}
\end{table}

\begin{figure}[!htbp]
\centering\includegraphics[width=.96\linewidth,height=.34\textheight,keepaspectratio]{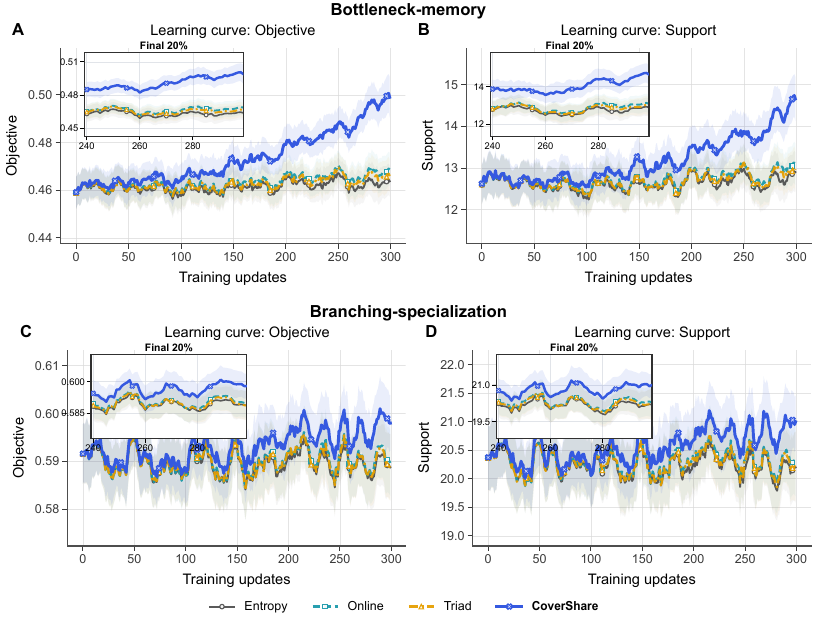}
\caption{Training dynamics in the controlled tasks. Panels A,B show bottleneck-memory and C,D show branching-specialization. Left and right columns show Objective and Support, respectively. Curves use a centered nine-update moving average across eight seeds; bands denote standard errors. Insets enlarge the final 20\% of updates.}\label{fig:controlled-learning}
\end{figure}

In bottleneck-memory, the learning curves separated as training progressed, with CoverShare continuing to improve during later updates. Branching-specialization showed smaller positive differences alongside similar fluctuations (Fig.~\ref{fig:controlled-learning}). This pattern is consistent with the structural difference between shared corridors and alternative branches. Section~\ref{sec:ablation-and-mechanism-analysis} examines credit components and changes in coverage overlap.

\noindent\textbf{DoorKey.}
Mean Support was 27.89 for RND, 28.53 for ETD, and 29.01 for CoverShare; corresponding Objective values were 0.6369, 0.6415, and 0.6453. CoverShare covered 0.48 more states than ETD, with both metrics higher in seven of eight paired seeds. ETD achieved higher mean Support than TEE on all three tasks. With the auxiliary coefficient fixed at 0.1, final-window unweighted rewards per step ranged across tasks from \(1.6\times10^{-8}\) to \(1.1\times10^{-6}\) for TEE and 0.15 to 0.23 for ETD.

\FloatBarrier
\Needspace{3\baselineskip}
\subsubsection{Public benchmarks and original-protocol validation}\label{sec:public-protocol-validation}

\begin{figure}[!htbp]
\centering
\includegraphics[width=.96\linewidth,height=.25\textheight,keepaspectratio]{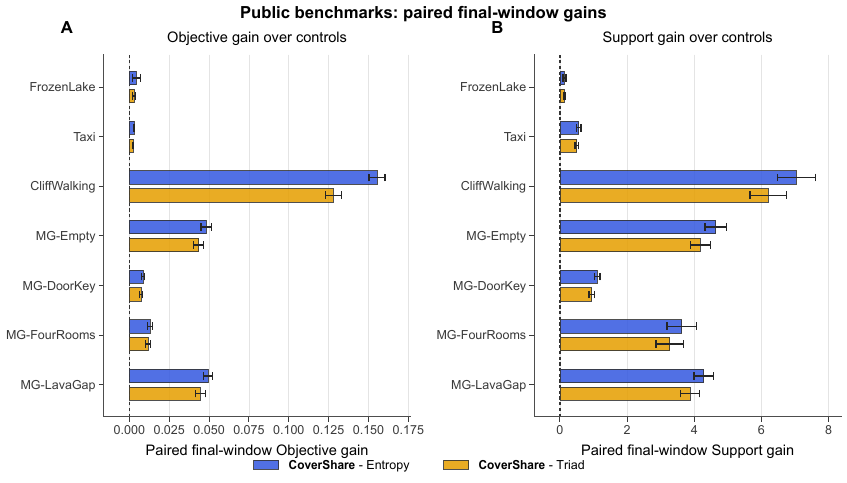}
\caption{Public-benchmark validation. Paired final-window gains of \ours{CoverShare} over Entropy and Triad are shown for Objective (A) and Support (B) across seven tasks; whiskers denote unadjusted bootstrap 95\% confidence intervals.}
\label{fig:9}
\end{figure}

\noindent\textbf{Public benchmarks.}\label{sec:public-discrete-state-benchmarks}
CoverShare achieved higher mean final-window Objective and Support than Entropy and Triad on all seven public tasks (Fig.~\ref{fig:9}). Giving tasks equal weight, mean Objective increased by 0.0403 over Entropy, and the mean relative Support gain was 16.29\% (exact \(p=0.0078\) for both). Coverage gains varied by task, including 7.038 additional states in CliffWalking and 0.134 in FrozenLake. Table~\ref{tab:7} provides all taskwise estimates. Holm correction across seven tasks within each metric gave taskwise \(p=0.0547\), above the 0.05 significance threshold.

\begin{table}[!htbp]
\caption{Final-window performance on public benchmarks.}\label{tab:7}
\begin{adjustwidth}{-\extralength}{0cm}
\setlength{\linewidth}{\fulllength}

\centering\fontsize{8}{10}\selectfont\setlength{\tabcolsep}{3pt}\renewcommand{\arraystretch}{1.12}
\textbf{(A) Objective}\par\smallskip
\begin{tabular*}{\fulllength}{@{\extracolsep{\fill}}lrrrrr@{}}
\toprule
Task & Entropy & Triad & \ours{CoverShare} & \(\Delta\) {[}95\% CI{]} & Holm-adjusted \(p\) \\
\midrule
FrozenLake-v1 & 0.5546 & 0.5560 & \ours{0.5589} & +0.0044 {[}0.0021, 0.0066{]} & 0.0547 \\
Taxi-v3 & 0.5256 & 0.5260 & \ours{0.5285} & +0.0029 {[}0.0026, 0.0032{]} & 0.0547 \\
CliffWalking-v0 & 0.5245 & 0.5518 & \ours{0.6798} & +0.1553 {[}0.1507, 0.1601{]} & 0.0547 \\
MiniGrid-Empty-8x8-v0 & 0.5004 & 0.5053 & \ours{0.5487} & +0.0483 {[}0.0451, 0.0513{]} & 0.0547 \\
MiniGrid-DoorKey-8x8-v0 & 0.6367 & 0.6380 & \ours{0.6453} & +0.0086 {[}0.0078, 0.0092{]} & 0.0547 \\
MiniGrid-FourRooms-v0 & 0.4945 & 0.4958 & \ours{0.5076} & +0.0131 {[}0.0117, 0.0146{]} & 0.0547 \\
MiniGrid-LavaGapS7-v0 & 0.5406 & 0.5454 & \ours{0.5899} & +0.0493 {[}0.0465, 0.0520{]} & 0.0547 \\
\bottomrule
\end{tabular*}
\par\medskip\textbf{(B) Support}\par\smallskip
\begin{tabular*}{\fulllength}{@{\extracolsep{\fill}}lrrrrr@{}}
\toprule
Task & Entropy & Triad & \ours{CoverShare} & \(\Delta\) {[}95\% CI{]} & Holm-adjusted \(p\) \\
\midrule
FrozenLake-v1 & 9.23 & 9.23 & \ours{9.36} & +0.134 {[}0.095, 0.175{]} & 0.0547 \\
Taxi-v3 & 34.38 & 34.45 & \ours{34.94} & +0.564 {[}0.503, 0.620{]} & 0.0547 \\
CliffWalking-v0 & 14.04 & 14.86 & \ours{21.07} & +7.038 {[}6.533, 7.581{]} & 0.0547 \\
MiniGrid-Empty-8x8-v0 & 18.65 & 19.10 & \ours{23.28} & +4.637 {[}4.343, 4.949{]} & 0.0547 \\
MiniGrid-DoorKey-8x8-v0 & 27.89 & 28.06 & \ours{29.01} & +1.115 {[}1.041, 1.192{]} & 0.0547 \\
MiniGrid-FourRooms-v0 & 40.44 & 40.80 & \ours{44.07} & +3.627 {[}3.237, 4.032{]} & 0.0547 \\
MiniGrid-LavaGapS7-v0 & 18.67 & 19.07 & \ours{22.95} & +4.281 {[}3.986, 4.545{]} & 0.0547 \\
\bottomrule
\end{tabular*}
\end{adjustwidth}
\tablefooter{Results are based on eight matched seeds. \(\Delta\) is \ours{CoverShare} minus Entropy with unadjusted bootstrap 95\% confidence intervals. Holm-adjusted \(p\) values correct across seven tasks within each metric.}
\end{table}

\begin{figure}[!htbp]
\centering
\includegraphics[width=.96\linewidth,height=.25\textheight,keepaspectratio]{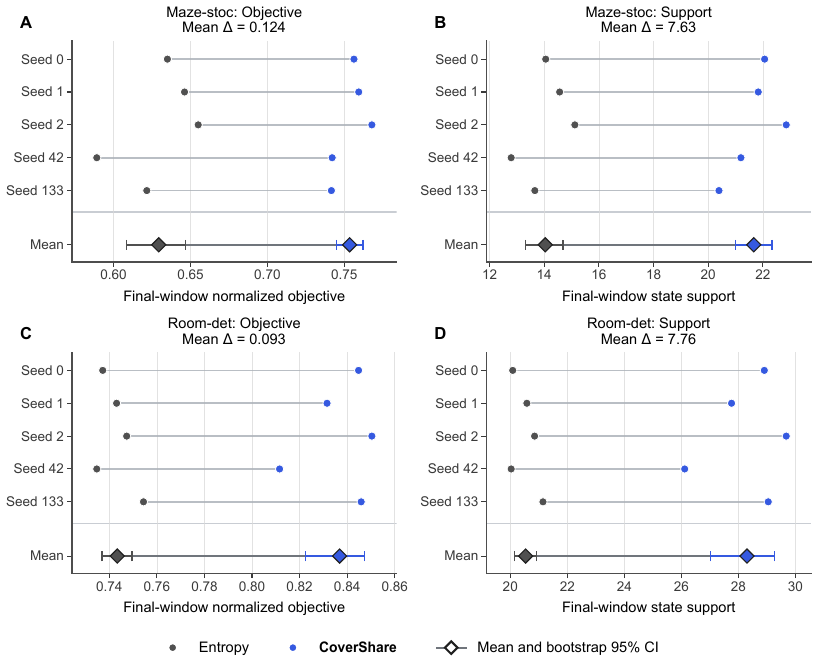}
\caption{Original-protocol validation. Paired final-window Entropy and \ours{CoverShare} results are shown for five seeds in stochastic Maze (A,B) and deterministic Room (C,D); Mean rows include bootstrap 95\% confidence intervals.}
\label{fig:7}
\end{figure}

\noindent\textbf{Original protocol.}\label{sec:original-protocol-replication}
Under the original PGPSE training configurations, both final-window metrics favored CoverShare over Entropy in all five paired seeds in Room and Maze (Fig.~\ref{fig:7}). Mean Objective and Support gains were 0.124 and 7.63 in Maze, and 0.093 and 7.76 in Room. With five pairs, the smallest attainable exact two-sided sign-flip \(p\) value is 0.0625, above the 0.05 significance threshold.

\FloatBarrier
\Needspace{3\baselineskip}
\subsubsection{Robustness under stochastic transitions}\label{sec:robustness-under-stochastic-transitions}

\begin{figure}[!htbp]
\centering
\includegraphics[width=.96\linewidth,height=.34\textheight,keepaspectratio]{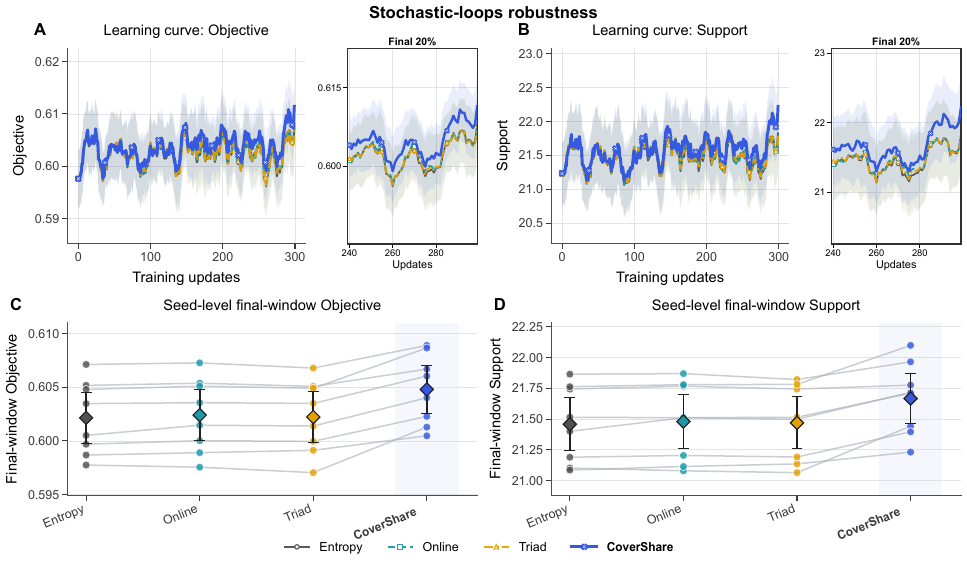}
\caption{Stochastic-loops robustness across eight matched seeds. (A,B) Mean learning curves with standard-error bands and final-window insets. (C,D) Paired final-window estimates with mean 95\% confidence intervals.}
\label{fig:8}
\end{figure}

In stochastic-loops, CoverShare increased final-window Objective and Support by 0.0026 and 0.209 relative to Entropy and by 0.0026 and 0.198 relative to Triad. All eight seed pairs favored CoverShare on both metrics against both baselines (Fig.~\ref{fig:8}, Holm-adjusted \(p\leq0.0234\) for each comparison).

Under stochastic transitions, both coverage gains persisted and were smaller than those observed in bottleneck-memory and branching-specialization.

\FloatBarrier
\Needspace{3\baselineskip}
\subsection{Ablation and mechanism analysis}\label{sec:ablation-and-mechanism-analysis}

The preceding comparisons established coverage performance in orchard scouting, controlled tasks, and public benchmarks. We now use paired ablations to examine credit components, allocation targets, and contribution-policy correspondence.

Which coverage signal best guides reward allocation, and how much does it matter which policy receives each weight? On both primary tasks, we varied the credit components, the definition of credit, the reward being weighted, and the correspondence between weights and policies. All variants used the same policy architecture, training budget, and eight matched seeds.

\FloatBarrier
\Needspace{3\baselineskip}
\subsubsection{Component ablation of marginal coverage credit}\label{sec:component-ablation-of-marginal-coverage-credit}

We compared four variants in the component ablation. They were Triad, LOO-only, Specialization-only, and CoverShare. Triad disabled marginal credit. LOO-only retained leave-one-policy-out support loss, whereas Specialization-only retained state-owner specialization. CoverShare combined both signals.

\begin{figure}[!htbp]
\centering
\includegraphics[width=.96\linewidth,height=.34\textheight,keepaspectratio]{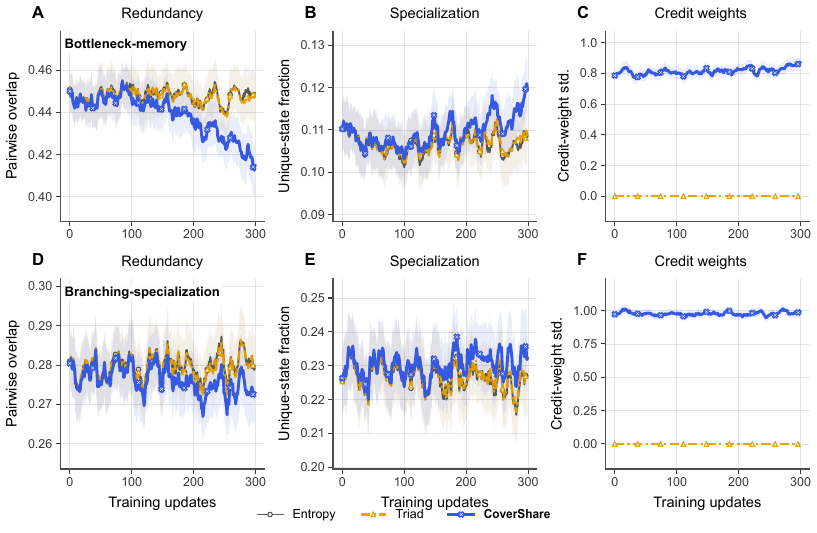}
\caption{Behavioral diagnostics for Entropy, Triad, and \ours{CoverShare} in bottleneck-memory (A--C) and branching-specialization (D--F): pairwise overlap, unique-state fraction, and credit-weight standard deviation. Curves show smoothed means with standard-error bands.}
\label{fig:5}
\end{figure}

\begin{table}[!htbp]
\caption{Component and mechanism-target ablation results.}\label{tab:4}
\begin{adjustwidth}{-\extralength}{0cm}
\setlength{\linewidth}{\fulllength}

\centering\fontsize{8}{10}\selectfont\setlength{\tabcolsep}{3pt}\renewcommand{\arraystretch}{1.12}
\begin{tabular*}{\fulllength}{@{\extracolsep{\fill}}lrrrr@{}}
\toprule
Environment / variant & Objective & Support & Pairwise overlap & Unique-state fraction \\
\midrule
BM / Triad & 0.4656 & 12.86 & 0.445 & 0.108 \\
BM / Specialization-only & 0.4682 & 12.97 & 0.444 & 0.108 \\
BM / LOO-only & 0.4866 & 13.83 & 0.427 & 0.113 \\
BM / Entropy-marginal & 0.4715 & 13.10 & 0.441 & 0.108 \\
BM / \ours{CoverShare} & \ours{0.4911} & \ours{14.04} & \ours{0.423} & \ours{0.114} \\
BM / CoverShare-on-entropy & 0.5621 & 18.16 & 0.352 & 0.136 \\
BS / Triad & 0.5899 & 20.26 & 0.280 & 0.225 \\
BS / Specialization-only & 0.5905 & 20.31 & 0.280 & 0.226 \\
BS / LOO-only & 0.5959 & 20.78 & 0.275 & 0.231 \\
BS / Entropy-marginal & 0.5911 & 20.35 & 0.279 & 0.226 \\
BS / \ours{CoverShare} & \ours{0.5963} & \ours{20.82} & \ours{0.274} & \ours{0.231} \\
BS / CoverShare-on-entropy & 0.6330 & 24.24 & 0.242 & 0.269 \\
\bottomrule
\end{tabular*}
\end{adjustwidth}
\tablefooter{Results are based on eight matched seeds. Entropy-marginal replaces LOO support loss with direct marginal-entropy credit; CoverShare-on-entropy applies coverage weights directly to team entropy. Results come from the matched ablation batch and may differ slightly from Table~\ref{tab:3}.}
\end{table}

LOO-only recovered most of the complete method's performance, whereas Specialization-only remained closer to Triad (Table~\ref{tab:4}). This pattern is consistent with the information captured by each signal. LOO support loss measures how much team support disappears when a policy trajectory is removed. State-owner specialization measures how many policies share a visited state but does not summarize full-trajectory contribution. CoverShare exceeded both single-component variants on final-window Objective and Support in both environments (Holm-adjusted \(p\leq0.0234\)). This result suggests that state-owner specialization provided a smaller refinement to the dominant LOO signal.

Coverage weights varied across policies and over time (Fig.~\ref{fig:5}). Compared with Triad, CoverShare reduced final-window pairwise overlap by 0.022 in bottleneck-memory and 0.006 in branching-specialization. The fraction of states visited by just one policy increased by approximately 0.006 in both tasks (Table~\ref{tab:4}).

\FloatBarrier
\Needspace{3\baselineskip}
\subsubsection{Credit definition and allocation target}\label{sec:credit-definition-and-allocation-target}

Removing one policy reduces state support by the number of states that policy alone visited. Marginal entropy also responds to changes in the frequency of visits to common states. With the same auxiliary-reward allocation rule, support loss yielded higher final-window Objective and Support than marginal entropy in all eight seed pairs on both controlled tasks (Holm-adjusted \(p=0.0078\)).

Applying the same coverage weights to team entropy instead of auxiliary rewards increased coverage on both primary tasks (Table~\ref{tab:4}). This variant, CoverShare-on-entropy, raised mean support from 14.04 to 18.16 states in bottleneck-memory and from 20.82 to 24.24 in branching-specialization. The corresponding gains in Objective were 0.0711 and 0.0367 (Holm-adjusted \(p=0.0078\) for each of the four comparisons). Direct weighting preserves the mean team score and may give coverage contributions greater influence on the policy gradient.

\FloatBarrier
\Needspace{3\baselineskip}
\subsubsection{Static and Reversed credit controls}\label{sec:static-and-reversed-credit-controls}

To distinguish non-uniform weighting from correct contribution alignment, we introduced two matched controls. Static credit used fixed non-uniform weights without trajectory-conditioned credit. Reversed credit performed the same dynamic marginal-credit calculation as CoverShare but assigned the resulting weights to policies in deterministic reverse order. The latter preserves credit computation and dynamic variation while breaking the correspondence between a contribution and the policy that produced it. None of the credit mappings added trainable parameters.

\begin{figure}[!htbp]
\centering
\includegraphics[width=.96\linewidth,height=.34\textheight,keepaspectratio]{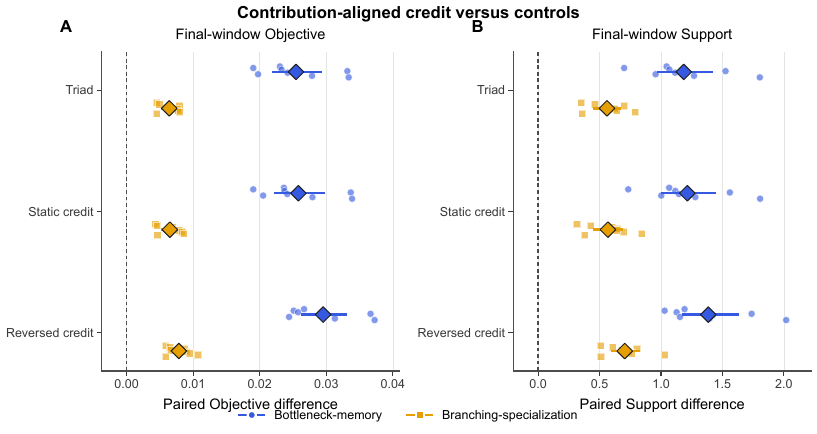}
\caption{Credit-alignment controls across eight seeds. Paired CoverShare-minus-control differences are shown for Objective (A) and Support (B); diamonds and whiskers denote means and bootstrap 95\% confidence intervals.}
\label{fig:6}
\end{figure}

Across both environments, all CoverShare-minus-control differences were positive for Objective and Support. Comparisons against Triad, Static credit, and Reversed credit remained significant after Holm correction (adjusted \(p=0.0234\) for each metric; Fig.~\ref{fig:6}). Static and Reversed credit remained close to Triad (Table~\ref{tab:5}). Neither non-uniformity alone nor reversed assignment reproduced the gain. Thus, contribution-policy correspondence was necessary among the tested controls.

\begin{table}[!htbp]
\caption{Final-window credit-control ablation results.}\label{tab:5}
\begin{adjustwidth}{-\extralength}{0cm}
\setlength{\linewidth}{\fulllength}

\centering\fontsize{8}{10}\selectfont\setlength{\tabcolsep}{3pt}\renewcommand{\arraystretch}{1.12}
\begin{tabular}{@{}llrr@{}}
\toprule
Env. & Variant & Objective & Support \\
\midrule
BM & \ours{CoverShare} & \ours{0.4911} & \ours{14.04} \\
BM & Static credit & 0.4652 & 12.83 \\
BM & Reversed credit & 0.4615 & 12.66 \\
BM & Triad & 0.4656 & 12.86 \\
BS & \ours{CoverShare} & \ours{0.5963} & \ours{20.82} \\
BS & Static credit & 0.5898 & 20.25 \\
BS & Reversed credit & 0.5885 & 20.11 \\
BS & Triad & 0.5899 & 20.26 \\
\bottomrule
\end{tabular}
\end{adjustwidth}
\tablefooter{Results are based on eight seeds.}
\end{table}

\begin{table}[!htbp]
\caption{Paired final-window differences between CoverShare and credit controls.}\label{tab:6}
\begin{adjustwidth}{-\extralength}{0cm}
\setlength{\linewidth}{\fulllength}

\centering\fontsize{8}{10}\selectfont\setlength{\tabcolsep}{3pt}\renewcommand{\arraystretch}{1.12}
\begin{tabular*}{\fulllength}{@{\extracolsep{\fill}}llrrrr@{}}
\toprule
Env. & Control & \(\Delta\) Objective & \(\Delta\) Support & Paired rank-biserial & Holm-adjusted \(p\) \\
\midrule
BM & Triad & +0.0255 & +1.185 & 1.00 / 1.00 & 0.0234 / 0.0234 \\
BM & Static credit & +0.0258 & +1.214 & 1.00 / 1.00 & 0.0234 / 0.0234 \\
BM & Reversed credit & +0.0295 & +1.384 & 1.00 / 1.00 & 0.0234 / 0.0234 \\
BS & Triad & +0.0064 & +0.560 & 1.00 / 1.00 & 0.0234 / 0.0234 \\
BS & Static credit & +0.0065 & +0.568 & 1.00 / 1.00 & 0.0234 / 0.0234 \\
BS & Reversed credit & +0.0079 & +0.705 & 1.00 / 1.00 & 0.0234 / 0.0234 \\
\bottomrule
\end{tabular*}
\end{adjustwidth}
\tablefooter{Differences are \ours{CoverShare} minus the corresponding control. Effect sizes and Holm-adjusted \(p\) values are reported as Objective/Support.}
\end{table}

\FloatBarrier
\Needspace{3\baselineskip}
\subsubsection{Effect of fixed-budget arbitration}\label{sec:negative-controls-and-attribution-boundary}

Across eight matched seeds, Triad's fixed-budget arbitration produced slightly lower final-window means than adding the auxiliary rewards directly. In bottleneck-memory, Objective was lower by 0.00126 (\(p=0.0078\)) and Support by 0.059 states (\(p=0.03125\)). In branching-specialization, Objective was lower by 0.00044 (\(p=0.0234\)) and Support by 0.043 states (\(p=0.0078\)). All four \(p\) values are Holm-adjusted.

\FloatBarrier
\Needspace{3\baselineskip}
\subsection{Hyperparameter sensitivity analysis}\label{sec:hyperparameter-sensitivity-analysis}

We examined sensitivity to three CoverShare parameters in bottleneck-memory: credit temperature, leave-one-policy-out coverage coefficient, and state-owner specialization coefficient. Each parameter varied separately while the others remained at their default values. Table~\ref{tab:8} reports Objective AUC over the complete run and final-window coverage metrics.

\begin{table}[!htbp]
\caption{One-factor credit sensitivity in bottleneck-memory.}\label{tab:8}
\begin{adjustwidth}{-\extralength}{0cm}
\setlength{\linewidth}{\fulllength}

\centering\fontsize{8}{10}\selectfont\setlength{\tabcolsep}{3pt}\renewcommand{\arraystretch}{1.12}
\begin{tabular*}{\fulllength}{@{\extracolsep{\fill}}llrrrrr@{}}
\toprule
Parameter & Value & Objective AUC & Final Objective & Final Support & Overlap & Unique-state fraction \\
\midrule
Temperature & 0.10 & 0.4777 (0.0026) & 0.5036 (0.0068) & 14.49 (0.33) & 0.4163 (0.0103) & 0.1151 (0.0025) \\
Temperature & 0.25 & 0.4748 (0.0027) & 0.4955 (0.0053) & 14.14 (0.27) & 0.4237 (0.0100) & 0.1136 (0.0018) \\
Temperature & 0.50* & 0.4707 (0.0019) & 0.4837 (0.0044) & 13.58 (0.24) & 0.4345 (0.0082) & 0.1102 (0.0020) \\
Temperature & 1.00 & 0.4661 (0.0012) & 0.4736 (0.0038) & 13.14 (0.16) & 0.4427 (0.0049) & 0.1072 (0.0006) \\
Temperature & 2.00 & 0.4637 (0.0010) & 0.4680 (0.0032) & 12.91 (0.13) & 0.4472 (0.0043) & 0.1063 (0.0012) \\
LOO coeff. & 0.00 & 0.4626 (0.0007) & 0.4656 (0.0025) & 12.81 (0.09) & 0.4488 (0.0035) & 0.1066 (0.0009) \\
LOO coeff. & 0.25 & 0.4649 (0.0011) & 0.4706 (0.0037) & 13.02 (0.15) & 0.4453 (0.0045) & 0.1066 (0.0009) \\
LOO coeff. & 0.50 & 0.4672 (0.0015) & 0.4759 (0.0041) & 13.24 (0.17) & 0.4409 (0.0063) & 0.1081 (0.0014) \\
LOO coeff. & 1.00* & 0.4707 (0.0019) & 0.4837 (0.0044) & 13.58 (0.24) & 0.4345 (0.0082) & 0.1102 (0.0020) \\
LOO coeff. & 2.00 & 0.4742 (0.0027) & 0.4936 (0.0054) & 14.07 (0.30) & 0.4253 (0.0097) & 0.1133 (0.0020) \\
Spec. coeff. & 0.00 & 0.4694 (0.0019) & 0.4807 (0.0043) & 13.46 (0.26) & 0.4373 (0.0084) & 0.1099 (0.0023) \\
Spec. coeff. & 0.25 & 0.4701 (0.0019) & 0.4824 (0.0040) & 13.53 (0.21) & 0.4355 (0.0077) & 0.1104 (0.0017) \\
Spec. coeff. & 0.50* & 0.4707 (0.0019) & 0.4837 (0.0044) & 13.58 (0.24) & 0.4345 (0.0082) & 0.1102 (0.0020) \\
Spec. coeff. & 1.00 & 0.4716 (0.0022) & 0.4867 (0.0045) & 13.73 (0.24) & 0.4315 (0.0089) & 0.1109 (0.0017) \\
Spec. coeff. & 2.00 & 0.4737 (0.0025) & 0.4922 (0.0048) & 14.00 (0.29) & 0.4264 (0.0098) & 0.1125 (0.0019) \\
\bottomrule
\end{tabular*}
\end{adjustwidth}
\tablefooter{Experiments used an RTX 3060. Values are mean (standard deviation [SD]) across five matched seeds. Asterisks mark default settings; the same default run is repeated within each parameter block.}
\end{table}

Credit temperature showed the clearest descriptive pattern. Within the tested range, lower temperatures were associated with higher Objective AUC, final Objective, and final Support. They were also associated with lower policy overlap and a higher unique-state fraction. This monotonic pattern is consistent with sharper allocation favoring policies that provide irreplaceable coverage.

Increasing the leave-one-policy-out coefficient from 0 to 2 raised mean final-window Support from 12.81 to 14.07. Varying the state-owner specialization coefficient over the same range raised Support from 13.46 to 14.00. The larger change for the leave-one-policy-out term agrees with the component ablation in Section~\ref{sec:component-ablation-of-marginal-coverage-credit}.

\FloatBarrier
\Needspace{3\baselineskip}
\subsection{Computational cost analysis}\label{sec:computational-cost-analysis}

We profiled Entropy, Triad, and CoverShare on one NVIDIA GeForce RTX 3060 (12 GB) under identical settings (Table~\ref{tab:9}). Each variant used five matched seeds and 300 updates. Every update contained eight rollout groups with a horizon of 20 across six parallel policies. The first 10 updates of each run were excluded as warm-up. Training time included rollout collection, exploration updates, and policy optimization. It excluded metric serialization, progress rendering, and checkpoint writing.

\begin{table}[!htbp]
\caption{Computational cost of exploration methods.}\label{tab:9}
\begin{adjustwidth}{-\extralength}{0cm}
\setlength{\linewidth}{\fulllength}

\centering\fontsize{8}{10}\selectfont\setlength{\tabcolsep}{3pt}\renewcommand{\arraystretch}{1.12}
\begin{tabular*}{\fulllength}{@{\extracolsep{\fill}}lrrrr@{}}
\toprule
Metric & Entropy & Triad & \ours{CoverShare} & \ours{CoverShare} vs Triad \\
\midrule
Time/update (s) & 1.2847 (0.0258) & 2.9392 (0.0253) & \ours{3.5721 (0.0306)} & +21.53\% \\
Agent throughput (steps/s) & 747.5 (15.3) & 326.6 (2.8) & \ours{268.8 (2.3)} & -17.72\% \\
Peak allocated memory (MB) & 23.76 & 47.57 & \ours{47.57} & $<0.01\%$ \\
Peak reserved memory (MB) & 30.00 & 54.00 & \ours{54.00} & No change \\
Training-time parameters & 275,736 & 1,400,391 & \ours{1,400,391} & No change \\
Deployed policy parameters & 275,736 & 275,736 & \ours{275,736} & No change \\
Policy-only inference (ms) & 1.5963 (0.0166) & 1.4874 (0.0248) & \ours{1.5856 (0.0234)} & Same model \\
\bottomrule
\end{tabular*}
\end{adjustwidth}
\tablefooter{Experiments used an RTX 3060. Timing, throughput, and latency are mean (SD) across five matched seeds. Memory and parameter rows report peak values or counts. \ours{CoverShare} versus Triad isolates marginal-credit overhead.}
\end{table}

Relative to Entropy, time per training update increased by 128.79\% for Triad and 178.05\% for CoverShare. Adding marginal credit to the same exploration stack increased time per update by 21.53\% and reduced throughput by 17.72\% relative to Triad. Both variants had the same trainable parameter count and reached the same peak allocated memory of 47.57 MB (Table~\ref{tab:9}).

At deployment, all three variants run the same policy architecture with 275,736 parameters. Measured policy-only inference latency ranged from 1.49 to 1.60 ms.

\FloatBarrier
\Needspace{3\baselineskip}
\subsection{Scope and limitations}\label{sec:agricultural-scope}

The agricultural evaluation uses road and visibility annotations from one persimmon site. Coverage records an opportunity to view a row segment. In the orchard, canopy occlusion, illumination, and viewing distance affect the assessment of leaf color and the measurement of leaf area and internode length. A field trial could record which segments yield observations usable for each trait and compare those records with the predicted route coverage.

\FloatBarrier
\Needspace{5\baselineskip}
\section{Conclusions}\label{sec:conclusions}

CoverShare allocates auxiliary rewards according to each policy's marginal contribution to team state coverage. Routes can therefore share an approach and still earn different credit for the areas they subsequently reach. In the persimmon orchard simulation, distinct policies provided observation opportunities for more row segments under a short inspection budget. Policies trained with 40 actions per route continued to reach new units when given 80 or 120 actions. Controlled ablations identified leave-one-policy-out coverage as the main source of credit, and the seven-task public benchmark comparison supported its contribution to state coverage.

Future work should evaluate direct marginal-credit allocation to team state entropy across larger policy ensembles, high-dimensional observations, continuous actions, and actor--critic backbones.

\FloatBarrier
\authorcontributions{Junhao Cao: Writing -- review \& editing, Writing -- original draft, Visualization, Validation, Software, Resources, Project administration, Methodology. Tao Wang: Software. Zipeng Xie: Data curation, Conceptualization. Hongyi Xia: Methodology, Data curation. Jianian Wu: Supervision, Resources, Formal analysis. Lixia Huang: Funding acquisition.}

\conflictsofinterest{The authors declare that they have no known competing financial interests or personal relationships that could have appeared to influence the work reported in this study.}

\dataavailability{The CoverShare code repository is available at \url{https://github.com/CpoSheep/CoverShare}. The orchard evaluation uses an annotated road network, row visibility annotations, per-seed results for four inspection budgets, and example trajectories.}

\funding{This paper reports interim results from five projects. The Key Laboratory of Digital Agriculture (Project 05) is a 2023 research and innovation platform for higher education institutions in Hunan Province. A 2025 key research project of the Hunan Provincial Department of Education (25A0777) investigates a digital decision system for precision persimmon cultivation that integrates multiple models, edge intelligence, and low-altitude coordination. The other contributing projects are the Changde Engineering Technology Research Center for Digital and Intelligent Crop Quality and Safety (a 2023 Changde science and technology innovation platform), the Smart Internet of Things Research Center (a 2021 research platform at Hunan Applied Technology University, Project 07), and the Smart Agriculture Research and Innovation Team (a 2022 science and technology innovation team at the same university, Project 01).}

\begin{adjustwidth}{-\extralength}{0cm}
\bibliographystyle{unsrtnat}
\bibliography{references}
\end{adjustwidth}

\end{document}